\documentclass{article}

\usepackage[numbers]{natbib}
\usepackage[preprint]{neurips_2026}

\usepackage[utf8]{inputenc} 
\usepackage[T1]{fontenc}    
\usepackage{hyperref}       
\usepackage{url}            
\usepackage{booktabs}       
\usepackage{amsfonts}       
\usepackage{nicefrac}       
\usepackage{microtype}      
\usepackage{xcolor}         
\usepackage{graphicx}
\usepackage{amsmath} 
\usepackage{tikz}
\usepackage{subcaption}
\usepackage[ruled, linesnumbered]{algorithm2e}
\usepackage{tikz}
\usepackage{tabularx}
\usepackage{float}
\usepackage{ragged2e}
\newcolumntype{L}{>{\RaggedRight\arraybackslash}X}
\newcommand{\seq}{\stackrel{\mathcal S}{=}} 
\title{Generative Anchored Fields: \\ Endpoint Learning for Generative Modeling}

\author{%
  Deressa Wodajo Deressa\textsuperscript{*}, 
  Hannes Mareen, 
  Peter Lambert, 
  Glenn Van Wallendael \\
  Ghent University -- imec, IDLab \\
  Department of Electronics and Information Systems, Gent, Belgium \\
  \textsuperscript{*}\texttt{deressawodajo.deressa@ugent.be}, 
  \texttt{firstname.lastname@ugent.be} \\
  \url{https://media.idlab.ugent.be}
}

\begin{document}

\maketitle

\begin{abstract}
  We present Generative Anchored Fields (GAF), a generative model that learns independent endpoint predictors, \(J\) (noise) and \(K\) (data), from any point on a linear bridge. Instead of learning the local path along a trajectory, as in diffusion or flow-based models, GAF is trained to identify the endpoint coordinates of a bridge state directly via an endpoint regression objective. Central to our approach is the decoupling of time-dependent feature learning from endpoint prediction. Unlike the transport-based models that learn a monolithic time-conditioned predictor that maps noise to data via a unified field, GAF uses a shared trunk, \(\Phi\), which produces a time-conditioned feature bank and a pair of linear predictors, \(J\) and \(K\). This separation of feature learning from prediction enables \textit{Transport Algebra}: algebraic operations on multiple \(J\) and \(K\) heads for compositional sampling. Additionally, with class-specific \(K_n\) heads, where \(n\in\{1,...,N\}\) for \(N\) classes, GAF defines directed transport maps between a shared base noise distribution and multiple data domains, allowing controllable interpolation, multi-class composition, and semantic editing. This is achieved either directly on the predicted data coordinates (\(K\)) using (i) Iterative Endpoint Refinement (IER), a novel sampler that achieves high-quality generation in \(5-8\) forward steps, or (ii) on the derived velocity field \(v=K-J\). GAF achieves strong sample quality with an FID of 2.25 on ImageNet \(256\times256\) while treating compositional generation as an architectural primitive.
  Code available at \url{https://github.com/IDLabMedia/GAF}.
\end{abstract}

\section{Introduction}
Modern Deep Generative Modeling is largely dominated by methods that learn to reverse a data-to-noise trajectory, with key differences arising in the parametrization of this reverse process~\citep{song2021scorebased,albergo2023building,lipman_flow_2023}. These approaches can be broadly categorized into score-based models that estimate the gradient of the log density, denoising models that learn corruption targets such as noise or clean data, and velocity-based models that learn the velocity field. While diverse, these frameworks share a common reliance on learning local step-by-step dynamics along the trajectory. Moreover, generation typically requires simulating either ordinary differential equations (ODEs) or stochastic differential equations (SDEs). 
\begin{figure}[!htbp]    
    \centering
    \includegraphics[width=0.99\linewidth]{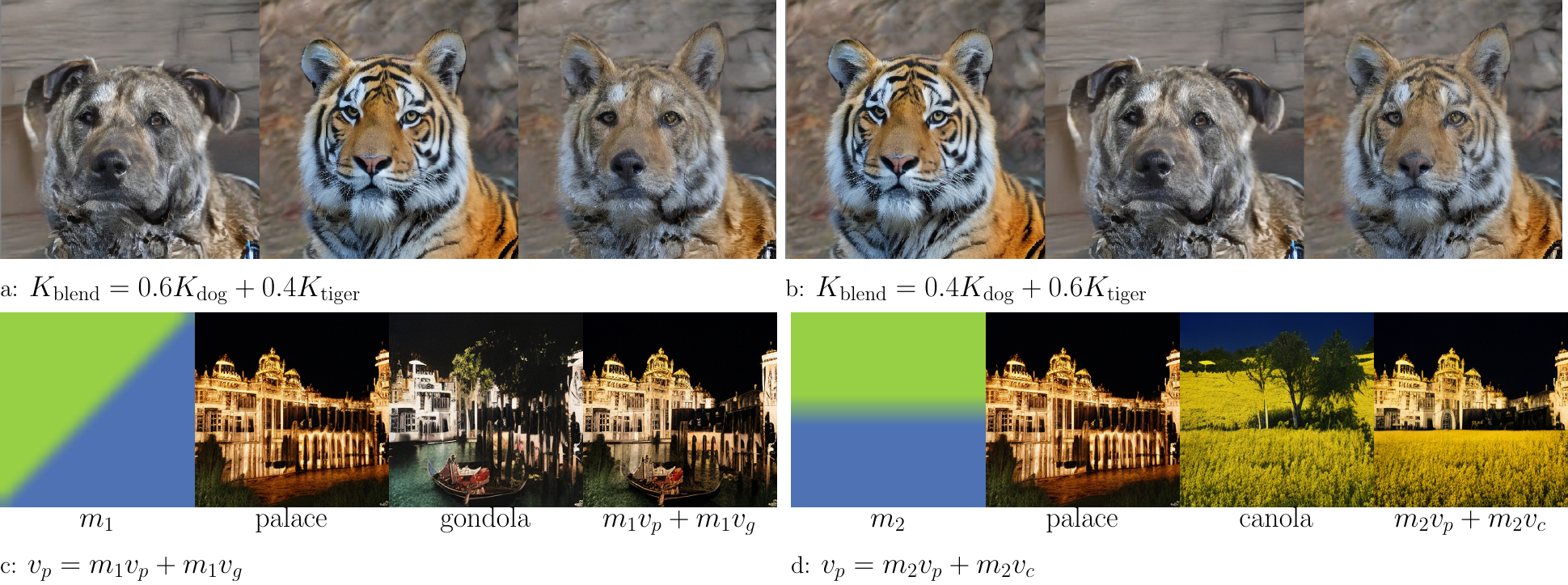}
    \caption{GAF enables compositional generation through algebraic combinations of endpoint predictors (\(K\)) and derived velocity fields (\(v=K-J\)). (a) Endpoint blending across AFHQ classes via weighted \(K\) heads. (b) Spatial scene composition on ImageNet via masked velocity fields, where \(m_1\) and \(m_2\) are spatial masks and \(v_p, v_g, v_c\) are class-specific velocity fields for palace, gondola, and canola.}
    \label{fig:teaser}
\end{figure}
The trajectory-based paradigm models generation through the evolution of states along a path. This raises a fundamental question: must we learn the local, step-by-step mechanism of the trajectory, or is it sufficient to learn only the endpoint pair that defines its origin and destination? 

We present Generative Anchored Fields (GAF), a framework that learns the endpoint coordinates of a linear interpolation (bridge) between a noise sample and a data sample. Unlike existing approaches that use a single trajectory or score predictor, GAF is trained to recover the bridge endpoints directly via coordinate learning. GAF uses two important units, the trunk and two endpoint predictors. The trunk's task is to learn the important features of the bridge between two initial and final probability distributions. Each predictor projects the trunk features to one of the bridge's endpoint coordinates. These linear head predictors have opposing objectives: one is anchored to a noise distribution and the other to the data manifold, and they act as simple projections that map the trunk features back to noise and data coordinates, respectively. 

The factorization of feature learning via trunk, and endpoint prediction via the linear heads make GAF inherently modular. For example, additional independent endpoint predictors \(J_n\) and \(K_n\), where \(n\in\{1,...,N\}\) for \(N\) classes or modalities, can be easily added to the same trunk without changing the underlying backbone. Since the heads operate on the same trunk and are linear in nature, they can be algebraically combined. We term this capability \textbf{Transport Algebra}: the ability to perform algebraic manipulations directly on the model's learned components to achieve precise semantic control.

Our proposed model has the following contributions:

\textbf{Coordinate-based Generative Modeling.} Instead of learning a score, noise, or velocity field, GAF learns the explicit coordinates of the transport endpoints. 

\textbf{Transport Algebra.} GAF introduces transport algebra to generative modeling: the ability to perform arithmetic operations directly on learned endpoint  predictors. This enables controllable test-time generation, such as interpolation between classes, switching manifolds mid-trajectory, or combining multiple classes directly on the \(K\) heads for novel compositional generation.

\textbf{Iterative Endpoint Refinement Sampler.} GAF has a native non-ODE sampler which iteratively refines endpoint estimates on a linear bridge, converging in \(5-8\) steps compared to \(80-250\) steps for Euler integration. 

\textbf{Inherent modularity and extensibility.} GAF's shared trunk with pluggable ``twin" or ``tuplet" heads is inherently modular. One can easily add additional \(K_{n}\) heads without altering the trunk.

\section{Related Work}
\label{related_work}
Modern generative models can be broadly categorized into denoising models that learn corruption targets such as noise or clean data, score-based models that estimate the gradient of the log density \(\nabla_{\mathbf{x}} \log p(\mathbf{x}_t)\), and velocity-based models that learn the velocity field. Score-based models, formalized through the lens of SDEs, established the score function as the key component needed to reverse a continuous-time forward diffusion process~\citep{song2021scorebased}. Concurrently, Denoising Diffusion Probabilistic Models (DDPMs)~\citep{pmlr-v37-sohl-dickstein15,ho_denoising_2020} simplified the transport objective to noise prediction \(\epsilon\). 

Most recently, the field has increasingly shifted towards a deterministic, continuous-time perspective rooted in ODEs and optimal transport~\citep{villani2009optimal,lipman_flow_2023,liu2022flow,lipman2024flowmatchingguidecode}. These models learn a velocity field \(v_{\theta}(\mathbf{x}_t,t)\) that governs a probability flow ODE. The development of simulation-free training objectives, such as Flow Matching~\citep{lipman_flow_2023} and Rectified Flows~\citep{liu2022flow}, has made this approach highly efficient. Stochastic Interpolants~\citep{JMLR:v26:23-1605} generalize this framework by introducing the bridge formulation between two distributions. The recent work of Negrel et al.~\citep{negrel2026multitask} extended this framework to operator-based interpolants, replacing scalar time variable with linear operators and learning two endpoint drifts, enabling zero-shot conditional generation across inpainting, fine-tuning and posterior sampling for lattice field theory, and trajectory planning from a single trained model. In the scalar-time setting their two endpoint drifts coincide with our twin predictors \(J\) and \(K\), and the derived velocity \(v=K-J\) recovers their probability flow. Our work was independently developed and is complementary in scope: their framework emphasizes operator-valued generalization and scientific applications, while, we focus on an architectural realization with shared trunk and class-specific linear heads, and the IER as a non-ODE alternative. While Consistency Models~\citep{song2023consistency} target few-step generation by distilling a pretrained diffusion model into a single-step predictor, GAF's IER achieves comparable step efficiency (\(5-8\) steps) with its native sampler without the need for a teacher model or distillation stage.

All these existing approaches, whether predicting score, noise, or velocity, typically learn a single time-conditioned predictor, and compositional control generally requires additional mechanisms. Recent work has explored alternative control mechanisms. For example, Compositional Visual Generation~\citep{liu2022compositional} defines two compositional operators (AND and NOT) for multi-concept generation, using a set of diffusion models to enable logical operation over concepts. Moreover, ControlNet~\citep{10377881} enables spatial control in text-to-image generation by locking the original diffusion model while training another network for conditional control with extra inputs like edges or depth. Various proposed methods such as Compositional Visual Generation~\citep{liu2022compositional}, and ControlNet~\citep{10377881} treat control as an add-on mechanism rather than an intrinsic architectural property. 

In contrast to existing generative models, GAF introduces an endpoint regression objective for transport modeling. Instead of a single monolithic local field, GAF learns two time-agnostic endpoint predictors \(J\) and \(K\). Compositional operations over class-specific heads follow as a consequence of this factorization.

\section {Generative Anchored Fields (GAF)}
\label{GAF}
\subsection{The GAF Model}
\label{GAF:Framework}
We introduce the Generative Anchored Fields model. Let \(\mathbf{z}_x, \mathbf{z}_y \in \mathbb{R}^{d}\) be samples from a data distribution \(\mathbf{z}_x\sim p_{\text{data}}(\mathbf{x})\) and a noise distribution \(\mathbf{z}_y\sim \mathcal{N}(0,I)\), respectively. 

The GAF model operates on linear bridges defined between these samples:
\begin{equation}\label{eq:bridge}
    \mathbf{x}_t = (1-t)\mathbf{z}_y + t\mathbf{z}_x, \quad t\in[0,1].
\end{equation}
The central task is for GAF to learn the two anchor endpoints \(\mathbf{z}_y\) and \(\mathbf{z}_x\) from a given point on the segment that they define. We call this task Coordinate Learning: the ability to identify the endpoints of a bridge from any point along its trajectory. The core of the model is a neural network, referred to as the trunk \(\Phi\), that processes the bridge \(\mathbf{x}_t\) and timestep \(t\) to produce a feature bank \(\mathbf{f}_t\):
\begin{equation}\label{eq:trunk}
    \mathbf{f}_t =  \Phi(\mathbf{x}_t,t). 
\end{equation}
Optionally, for class conditioning, we condition on \(\mathbf{c}\): \(\mathbf{f}_t =  \Phi(\mathbf{x}_t,t,\mathbf{c})\), where  \(\mathbf{c} \in {\mathbb{R}^d}\) is an optional conditioning vector (e.g., a class embedding). The trunk, \(\Phi(\mathbf{x}_t,t) \in \mathbb{R}^{d}\), provides the shared time-conditioned feature representation used by the two (twin) predictors, \(J\) and \(K\). We term \(J\) and \(K\) \textit{twins} to emphasize their symmetric roles: both are endpoint predictors that share the same feature from the trunk. Their velocity field \(v=K-J\) emerges from their difference. The trunk hides architectural complexity details from the twins and enables modularity (e.g., plugging different heads, conditioning, or label embeddings) and parameter sharing. 
The feature bank \(\mathbf{f}_t\) is then processed by a pair of twins \(J\) and \(K\), which regress the bridge \(\mathbf{x}_t\) towards its endpoints (i.e., noise \(\mathbf{z}_y\) and data \(\mathbf{z}_x\), respectively).
The twins are formally defined as:  
\begin{align}
    J &:= (1-t)\mathbf{x}_t + H_J(\mathbf{f}_t),\label{eq:J_definition}\\
    K &:= t \mathbf{x}_t + H_K(\mathbf{f}_t),\label{eq:K_definition}\\
    J, K &= \text{GAF}_{\theta}(\mathbf{x}_t,t).\label{eq:gaf_main_definition}
\end{align}
The two endpoints are predicted by a single forward pass through GAF, with shared trunk features feeding both heads~\eqref{eq:gaf_main_definition}. \(H_J\) and \(H_K\) are trained for separate tasks. \label{GAF:twins}We denote the residuals \(J_{\text{res}}:=H_J(\mathbf{f}_t)\) and \(K_{\text{res}}:=H_K(\mathbf{f}_t)\). We term our model ``Anchored Fields" because the predictors \(J\) and \(K\) are anchored to fixed endpoints: \(J\) targets the noise distribution, \(K\) targets the data manifold.

The endpoint predictors \(J\) and \(K\) move in opposite directions along the bridge. More specifically, as viewed from \(\mathbf{x}_t\), \(J\) is oriented toward the noise endpoint, and \(K\) toward the data endpoint. Their ideal boundary conditions are \(J=\mathbf{z}_y\) at \(t=0\), and \(K=\mathbf{z}_x\) at \(t=1\). To quantify the twins' motion relative to each other along the bridge, we compute the path's instantaneous velocity with respect to \(t\). For the linear bridge~\eqref{eq:bridge}, we have:
\begin{equation}
\frac{d\mathbf{x}_t}{dt}
= \frac{d}{dt}\big[(1-t)\mathbf{z}_y + t\mathbf{z}_x\big]
= \mathbf{z}_x - \mathbf{z}_y .
\label{eq:bridge_derivative}
\end{equation}
Because \(K\) and \(J\) are anchored to data and noise respectively, we define the learned velocity field \(v\) as the twins disagreement:
\begin{align}\label{eq:velocity}
    v(\mathbf{x}_t, t) = K - J.
\end{align}
This learned field then specifies an ODE that transports the noise sample \(\mathbf{z}_y\) to the data sample \(\mathbf{z}_x\):
\begin{equation}\label{eq:sampler}
    \frac{d\mathbf{{x}}_t}{dt} = v(\mathbf{x}_t,t).
\end{equation}
We term \(J\) and \(K\) \textit{twins} to emphasize their symmetric roles: both are endpoint predictors that share the same feature from the trunk. Their velocity field \(v=K-J\) emerges from their difference.

We generate new data samples by either (i) iteratively refining the endpoints via the novel IER algorithm (see Section~\ref{GAF:Sampling-IER}) or (ii) integrating the velocity field ODE from \(t=0\) to \(t=1\) along a linear or cosine time grid, using Euler (or Heun) integration~\citep{chen_neural_2018, 10.5555/3600270.3602196}. Importantly, because the heads are time-agnostic and anchored to the endpoints, we can sample with GAF without the traditional ODE/SDE simulations, enabling high-quality generation through direct coordinate refinement (see Section~\ref{GAF:Sampling}).

\subsubsection{GAF Training}
\label{GAF:training}
The model is trained using the linear endpoint's pair loss:
\begin{equation}\label{eq:endpoint_loss}
    \mathcal{L_{\text{pair}}}=\mathbb{E}_{\mathbf{z}_x, \mathbf{z}_y, t}[(1-t)\|J - \mathbf{z}_y\|_2^2 + t\|K - \mathbf{z}_x\|_2^2]
\end{equation}
At the population optimum of the endpoint pair loss (\({L_{\text{pair}}}\)), squared-error regression gives 
\[J=\mathbb{E}[\mathbf{z}_y\mid \mathbf{x}_t, t]\qquad K=\mathbb{E}[\mathbf{z}_x\mid \mathbf{x}_t, t]\] and therefore \[K-J = \mathbb{E}[\mathbf{z}_x - \mathbf{z}_y \mid \mathbf{x}_t, t]. \]
Thus, the endpoint pair objective recovers the marginal velocity field associated with the linear bridge.

\textbf{Residual penalty.} From \ref{eq:J_definition} and \ref{eq:K_definition}, achieving the ideal condition \(J=\mathbf{z}_y\) requires \(J_{\text{res}}|_{t=0}=0\). Likewise, \(K=\mathbf{z}_x\) requires \(K_{\text{res}}|_{t=1}=0\). While the primary endpoint loss \eqref{eq:endpoint_loss} implicitly encourages this conditions, we introduce an explicit residual regularization penalty to enforce them directly:
\begin{equation}\label{eq:residual_penalty}
    \mathcal{L_{\text{res}}}=\mathbb{E}_{\mathbf{z}_x, \mathbf{z}_y, t}[(1-t)\|J_{\text{res}}\|_2^2 + t\|K_{\text{res}}\|_2^2]
\end{equation}

\textbf{Residual-antisymmetry penalty (swap loss).} The endpoint residual penalty ensures \(J\rightarrow\mathbf{z}_y\) as \(t\rightarrow0\) and \(K\rightarrow\mathbf{z}_x\) as \(t\rightarrow1\), but it does not fully determine the intermediate behavior of GAF on the bridge. Since we want GAF to behave consistently across the entire bridge, we need a mechanism to do that. To this end, we swap the residuals \(J_{\text{res}}\) and \(K_{\text{res}}\) across the bridge and create antisymmetric relationship between them using the swap operator \(\mathcal{S}\):
\begin{equation}\label{eq:swap_function}
    \mathcal{S} : (\mathbf{z}_x,\mathbf{z}_y,t, c) \mapsto (\mathbf{z}_y,\mathbf{z}_x,1-t,c)
\end{equation}
Using our bridge formalism in \eqref{eq:bridge} and the swap operator in~\eqref{eq:swap_function}, we have:
\begin{align}
    \mathbf{x}_{1-t} &= t\mathbf{z}_x + (1-t)\mathbf{z}_y\label{eq:swap_bridge_equiv}
\end{align}
From \eqref{eq:bridge} and \eqref{eq:swap_bridge_equiv}, \(\mathbf{x}_t \seq \mathbf{x}_{1-t}\),  
i.e., the swap stays on the bridge, and only the direction of travel changes. Here, \(\seq\) means equality after applying the swap (namely, swapping endpoints and flipping time). Using our trunk from \eqref{eq:trunk} and swap function from \eqref{eq:swap_function}, the \textbf{endpoint-swap antisymmetric residuals} are defined as:
\begin{align}\label{eq:swapped_residuals}
    \tilde{J}_{\text{res}} := H_J(\Phi(\mathbf{x}_{1-t},1-t,c)),\quad
    \tilde{K}_{\text{res}} := H_K(\Phi(\mathbf{x}_{1-t},1-t,c)).
\end{align}
 A tilde above a quantity, as in~\eqref{eq:swapped_residuals}, indicates that it is evaluated in the swapped space. 
 Although GAF is trained with an endpoint regression, its emergent velocity field exhibits how it learns the transport dynamics (see Appendix~\ref{emergent_transport_dynamics}). As we have discussed earlier, in GAF, we do not train the velocity field directly, as in rectified flow (see Section~\ref{GAF:rectified_flow_comparison} for a comparison). Instead, the velocity simply emerges from the twins opposite movement. However, from~\eqref{eq:velocity_at_each_time} and~\eqref{eq:swap_velocity_at_each_time}, we can see that the emergent velocity field incorporates knowledge of the trajectory's endpoint.

\begin{figure}[t]
    \centering
    \includegraphics[width=\linewidth]{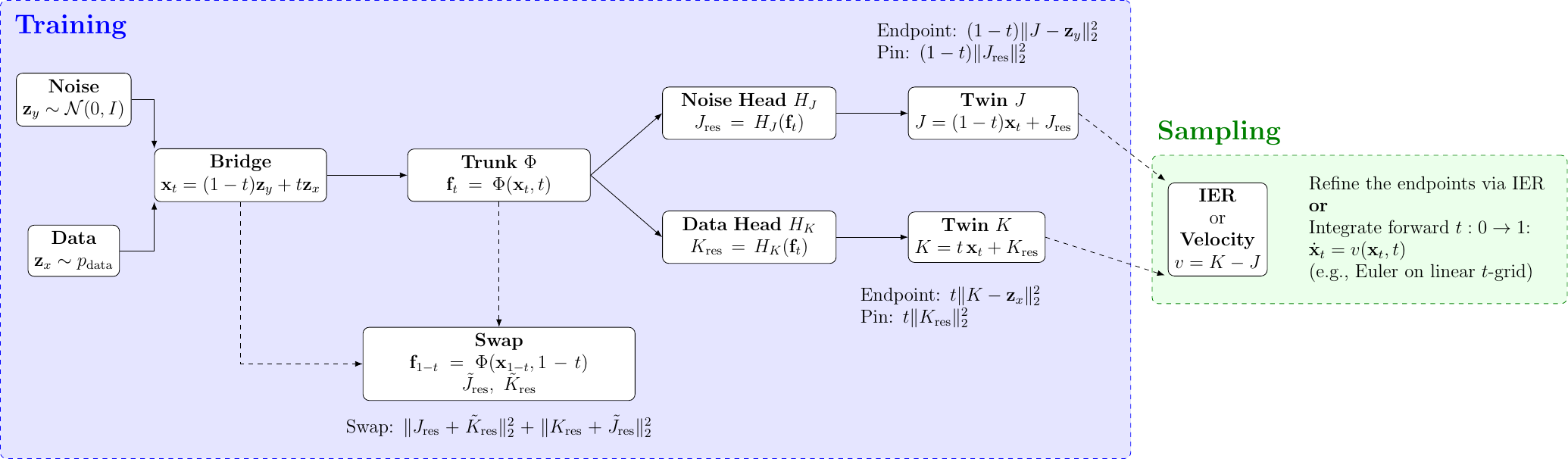}
    \caption{High-level overview of GAF architecture and its data flow.}
    \label{fig:data_flow}
\end{figure}

We then enforce a paired antisymmetric relationship between the original and swapped residuals: \(J_{\textbf{res}}\) pairs with \(\tilde{K}_{\text{res}}\) and \(K_{\textbf{res}}\) pairs with \(\tilde{J}_{\text{res}}\). Hence, we define the residual-antisymmetry penalty as:
\begin{align}
\mathcal{L}_{\text{swap}}
&= \|J_{\text{res}}+\tilde{K}_{\text{res}}\|_2^2 
 + \|K_{\text{res}}+\tilde{J}_{\text{res}}\|_2^2 .
\label{eq:endpoint_antisymmetric_loss}
\end{align}
From \eqref{eq:endpoint_loss}, \eqref{eq:residual_penalty}, and \eqref{eq:endpoint_antisymmetric_loss}, 
the GAF training loss is
\begin{equation}
\mathcal{L}_{\text{GAF}}
= \mathcal{L}_{\text{pair}}
+ \lambda_{\text{res}}\,\mathcal{L}_{\text{res}}
+ \lambda_{\text{swap}}\,\mathcal{L}_{\text{swap}},
\label{eq:GAF_loss}
\end{equation}
where \(\lambda_{\text{res}}\) and \(\lambda_{\text{swap}}\) are hyperparameters (e.g, both \(0.01\)). See the Ablation Study in Appendix~\ref{GAF:ablation_study}, where we isolate these loss functions to study their effect on the main GAF loss.\\
The overall architecture of GAF and its data flow is summarized in Figure~\ref{fig:data_flow}, illustrating how the training components (left) predict endpoints that drive sampling strategies (right) via IER or velocity integration.

\subsection{Transport Algebra}
\label{GAF:TA}
Transport Algebra organizes GAF’s endpoint pairs into an algebra of endpoint predictors, enabling composable operations that yield controlled data generation. Recall that GAF has two linear endpoints predictors \(J\) and \(K\). Because GAF is modular, the shared trunk can host multiple tuplets \(\{(J_{n},K_{n})\}_{n=1}^N\). This creates a natural algebra in which compositional operations can be performed (i) directly on the endpoint predictors by combining \(J_n\) and \(K_n\), and then applying endpoint refinement, and (ii) by using the emergent velocity field \(v_n=K_n-J_n\) and combining these fields algebraically. This ability of GAF to operate algebraically leads to a precise and semantically coherent method of data generation across multiple classes and domains.

\paragraph{Bridge and swap operators.}Using the swap operator \(\mathcal{S}\), we can define two additional bridge configurations as follows: 
\begin{align}
    \mathcal{S}_{\text{swap}} &: (\mathbf{z}_y,\mathbf{z}_x,t, c) \mapsto (\mathbf{z}_x,\mathbf{z}_y,t,c) && \text{(Swap endpoints)}\label{eq:pt1}\\
    \mathcal{S}_{\text{flip}} &: (\mathbf{z}_y,\mathbf{z}_x,t, c) \mapsto (\mathbf{z}_y,\mathbf{z}_x,1-t,c) && \text{(Flip time)}\label{eq:pt2}
\end{align}
From \eqref{eq:pt1} and \eqref{eq:pt2}, we get \(\mathbf{x}_t=(1-t)\mathbf{z}_x + t\mathbf{z}_y\). Just like the main bridge in \eqref{eq:swap_function}, the new swaps in \eqref{eq:pt1} and \eqref{eq:pt2} likewise remain on a bridge. Consequently, one can compose these operators to perform controlled generation at test time, using the resulting configurations to traverse or switch modalities (see Appendix~\ref{sec:evaluation-transport}). Figure~\ref{fig:swap_function} illustrates the four swap configurations. Additionally, we can have different \(J/K\) pairing topologies (see Appendix \ref{J_K_topologies})
\begin{figure}[t]
    \centering
    \includegraphics[width=0.99\linewidth]{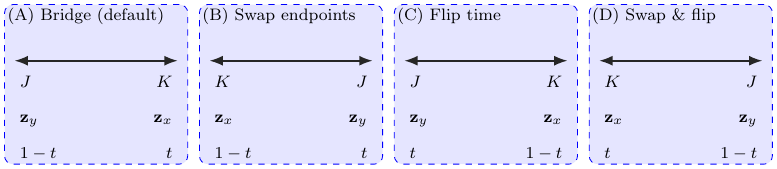}
    \caption{Visualization of the swap operator \(\mathcal{S}\) acting on a bridge. \textbf{(A)} The initial configuration of the bridge~\eqref{eq:bridge}. Subfigures \textbf{(B)}, \textbf{(C)}, and \textbf{(D)} illustrate the results of applying the operations described in equations~\eqref{eq:pt1},~\eqref{eq:pt2}, and~\eqref{eq:swap_function}, respectively.}
    \label{fig:swap_function}
\end{figure}
\paragraph{Transport algebra operations.} GAF's independent \(K\) heads enable compositional transport through endpoint operations (e.g, through IER or linear velocity operations). Let \(i,j \in \{1, \dots, N\}\).
\begin{enumerate}
    \item \label{GAF:single_class_transport} \textbf{Single-class transport.} Samples from class \(i\) can be generated by applying the IER algorithm (see Section \ref{GAF:Sampling-IER}) using the corresponding \(K_i\) head to iteratively refine the data endpoint from \(\mathbf{z}_y\sim \mathcal{N}(0,I)\). Alternatively, if velocity view is used and the noise endpoint \(J\) is shared, then \(v_i=K_i-J\) or if each modality has its own noise endpoint \(J_i\), then \(v_i = K_i-J_i\) operations can be used. Integrating \(v_i\) from noise (\(t=0\)) to data (\(t=1\)) generates samples from class \(i\).
    
    \item \label{GAF:cross_modality_transport} \textbf{Cross-modality transport (via \(J\)).}
    Move between data endpoints of two (or more) domains. To move from class \(i\) to class \(j\), we use \(J\) as a shared terminal between classes and \(K_{i\to j}\)=\(K_{i}\xrightarrow{J}K_j\) (IER) or \(v_{i\to j}=v_i\xrightarrow{-v_i}J\to v_j\) (ODE) (see Appendix \ref{cross_modality_transport}). 

    \item \label{GAF:direct_k_head_interpolation} \textbf{Direct \(K\)-head interpolation.} We can interpolate directly on \(K\) heads \(K_\alpha = (1-\alpha)K_i + \alpha K_j,\) or in the velocity space via weighted combinations \(v_\alpha = (1-\alpha)(K_i-J) + \alpha(K_j-J) = (1-\alpha)v_i + \alpha v_j,\) for \(\alpha\in[0,1]\). A simultaneous weighted combination of multiple \(K\) heads enables compositional generation, which is experimented on in Section~\ref{GAF:spatial_velocity_interpolation}. These operations demonstrate GAF's transport algebra where \(K\) independent heads act as linear operators that can be composed, interpolated, and combined to traverse the generative space. This property extends naturally to sequential domains.
For example, for video generation, we hypothesize that GAF can smoothly transition between temporal states while preserving semantic coherence through IER interpolation or velocity interpolation, which we will explore in future work (see Appendix~\ref{sec:evaluation-transport} for more examples on this).
    \item \label{GAF:multi_class_composition} \textbf{Multi-class composition (barycentric).} We can combine multiple \(K\) heads \(K_{\text{blend}}=\sum_{m} w_m K_m\) or velocity fields by linear vector operations \(v_{\text{blend}}=\sum_{m} w_m(K_m-J)\), where \(\sum_{m}w_m=1\). 
\end{enumerate}
By using \(J, K\), IER, \(v\), and the swap operator \(\mathcal{S}\) (which swaps \(J \leftrightarrow K\), \(z_y \leftrightarrow z_x\), and \(t \leftrightarrow 1-t\)), one can traverse modalities, switch modalities mid-generation, revert to the previous modality, chain back to the origin, or even use a specific generated modality as a bootstrap for a chained generation.  All of this is possible by switching endpoint heads and applying endpoint refinement, or by switching between the emergent velocity field \(v_i\), \(i \in \{1,\dots,N\}\), and applying simple vector operations. In all cases, the \(\mathcal{S}\) operator reconfigures bridges and directions on the fly. Detailed empirical analysis of Transport Algebra can be found in appendix \ref{sec:evaluation-transport}.

\subsection{Sampling} 
\label{GAF:Sampling}
GAF supports various sampling modes: IER (Section~\ref{GAF:Sampling-IER}), as well as other traditional ODE/SDE and hybrid-based samplings (see Appendix ~\ref{GAF:Sampling-ODE}, and ~\ref{GAF:Sampling-Hybrid}).

\subsubsection{Iterative Endpoint Refinement Sampling}\label{GAF:Sampling-IER}

Given the GAF factorization of its endpoint \(J\) and \(K\), we sample by iteratively refining endpoints on a linear bridge. In the forward direction, \(K\) is used to refine the data endpoint from the intermediate bridge points~\(\mathbf{x_t}\). In the reverse direction, \(J\) is used to refine the noise endpoint from intermediate bridge points. Both the bridge \(\mathbf{x_t}=(1-t)\mathbf{z_0}+t\mathbf{x_1}\) and the endpoints update (\(\mathbf{x}_1 \leftarrow (1-\alpha)\,\mathbf{x}_1 + \alpha\,K\) for data, \(\hat{\mathbf{z}} \leftarrow (1-\alpha)\,\hat{\mathbf{z}} + \alpha\,J\) for noise) are linear operations, so IER traverses straight paths between noise and data without ODE integration.

\textbf{Forward Generation (noise \(\to\) data).}
Starting from \(\mathbf{z}_y\sim \mathcal{N}(0,I)\), we initialize \(\mathbf{x_1}\leftarrow \mathbf{z_0}\), set \(\alpha\in[0,1]\), and iterate over \(t~\in [t_{\varepsilon},1-t_{\varepsilon}]\) (Algorithm \ref{alg:ier_forward}). At each step, we form \(\mathbf{x_t}=(1-t)\mathbf{z_0}+t\mathbf{x_1}\), predict \(K\), and update \(\mathbf{x_1}\) by a convex combination of the current data estimate and the prediction.

\textbf{Reverse Inversion (data \(\to\) noise).} 
Given the data anchor \(\mathbf{x_{\text{anchor}}}\), we initialize \(\hat{\mathbf{z}}\sim \mathcal{N}(0,I)\), set \(\alpha\in[0,1]\), and iterate over \(t~\in [t_{\varepsilon},1-t_{\varepsilon}]\) (Algorithm \ref{alg:ier_reverse}). At each step, we form \(\mathbf{x_t}=(1-t)\hat{\mathbf{z}}+t\mathbf{x_{\text{anchor}}}\), predict \(J\), and update \(\hat{\mathbf{z}}\) by a convex combination of the current noise estimate and the prediction.
\begin{center}
{
\scalebox{0.85}{
\begin{minipage}[t]{0.50\textwidth}
\vspace{0pt}
    \begin{algorithm}[H] 
        \caption{IER Forward (noise \(\rightarrow\) data; \(K\)-only)}
        \label{alg:ier_forward}
        \KwIn{noise \(\mathbf{z}_0\), steps \(S\), \(t_{\varepsilon}\), \(\alpha\), \(\mathrm{GAF}(\cdot,t)\)\vphantom{data anchor \(\mathbf{x}_{\mathrm{anchor}}\)}}
        \KwOut{\(\mathbf{x}_1\) (data estimate)}
        \(\mathbf{x}_1 \leftarrow \mathbf{z}_0\)\;
        \(t_1,\dots,t_S \leftarrow \mathrm{Linspace}(t_{\varepsilon},\,1-t_{\varepsilon},\,S)\)\;
        \For{\(s \leftarrow 1\) \KwTo \(S\)}{
            \(t \leftarrow t_s\)\;
            \(\mathbf{x}_t \leftarrow (1-t)\,\mathbf{z}_0 + t\,\mathbf{x}_1\)\;
            \((\_,K) \leftarrow \mathrm{GAF}(\mathbf{x}_t,t)\)\;
            \(\mathbf{x}_1 \leftarrow (1-\alpha)\,\mathbf{x}_1 + \alpha\,K\)\;
        }
        \Return{\(\mathbf{x}_1\)}\;
    \end{algorithm}
\end{minipage}%
\hspace{0.12\textwidth}%
\begin{minipage}[t]{0.50\textwidth}
\vspace{0pt}
    \begin{algorithm}[H]
        \caption{IER Reverse (data \(\rightarrow\) noise; \(J\)-only)}
        \label{alg:ier_reverse}
        \KwIn{data anchor \(\mathbf{x}_{\mathrm{anchor}}\), steps \(S\), \(t_{\varepsilon}\), \(\alpha\), \(\mathrm{GAF}(\cdot,t)\)}
        \KwOut{\(\hat{\mathbf{z}}\) (noise estimate)}
        \(\hat{\mathbf{z}} \sim \mathcal{N}(0,I)\)\;
        \(t_1,\dots,t_S \leftarrow \mathrm{Linspace}(1-t_{\varepsilon},\,t_{\varepsilon},\,S)\)\;
        \For{\(s \leftarrow 1\) \KwTo \(S\)}{
            \(t \leftarrow t_s\)\;
            \(\mathbf{x}_t \leftarrow (1-t)\,\hat{\mathbf{z}} + t\,\mathbf{x}_{\mathrm{anchor}}\)\;
            \((J,\_) \leftarrow \mathrm{GAF}(\mathbf{x}_t,t)\)\;
            \(\hat{\mathbf{z}} \leftarrow (1-\alpha)\,\hat{\mathbf{z}} + \alpha\,J\)\;
        }
        \Return{\(\hat{\mathbf{z}}\)}\;
    \end{algorithm}
\end{minipage}
}
}
\end{center}
\section{Experimental Evaluation}
\label{sec:evaluation}
In this section, we evaluate GAF for image generation, as proposed in Sections~\ref{GAF}.
Detailed description of the network architecture and experimental setup (incl. datasets and training procedure) can be found in Appendix \ref{detailed_arch_desc} and \ref{training_loss_desc}, respectively. 

\subsection{Multi-Class Generation (One \texorpdfstring{\(J\)}{J} and Multiple \texorpdfstring{\(K\)}{K}s)}
\label{sec:evaluation-multi}
\begin{figure}[t]
  \centering
  \begin{subfigure}[b]{\linewidth}
    \centering
    \includegraphics[width=0.99\linewidth]{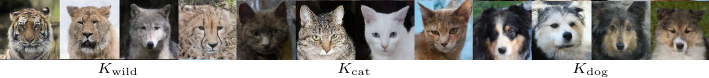}
    \caption{Samples from GAF trained with one \(J\) and three independent \(K\) heads, on three classes of the AFHQ dataset.}
    \label{fig:afhq}
  \end{subfigure}
  \vspace{0.5em}
  \begin{subfigure}[b]{\linewidth}
    \centering
    \includegraphics[width=0.99\linewidth]{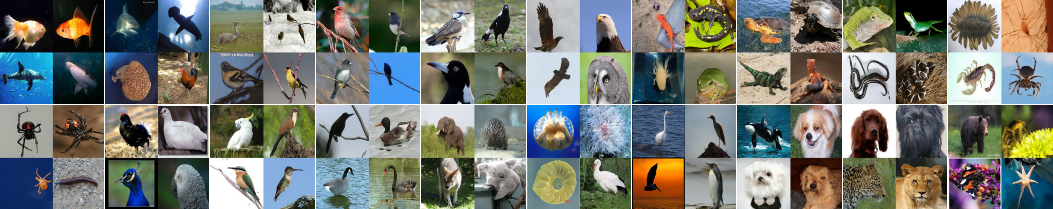}
    \caption{Samples from GAF trained with one \(J\) and \(1,000\) independent \(K\) heads on \(1,000\)-class ImageNet dataset.}
    \label{fig:imagenet_1000_class}
  \end{subfigure}
  \caption{Multi-class generation across two datasets. (a) AFHQ with three classes; (b) ImageNet with \(1{,}000\) classes.}
  \label{fig:multiclass_generation}
\end{figure}

This section evaluates GAF on multi-class generation, using three datasets. For single class generation, see Appendix \ref{sec:evaluation-single}. First, we utilize CIFAR-\(10\) (\(32\times32\) px) with ten object classes and one \(J\) head and ten \(K\) heads. Second, we utilize AFHQ (\(512\times512\) px) with three classes: \textit{cat}, \textit{dog}, and \textit{wild} animals, using one \(J\) head and three \(K\) heads. Third, we utilize ImageNet-1k with \(1,000\) classes, using one \(J\) head and \(1,000\) \(K\) heads. Figure~\ref{fig:afhq} and Figure~\ref{fig:imagenet_1000_class} show samples from each independent \(K\) head, each corresponding to one of the three or thousand classes within each dataset, respectively. This visually demonstrates that GAF scales to multi-class generation while maintaining high per-class quality.

\subsection{Evaluation}
\label{sec:qualitative_evaluation}
We analyze the effect of the number of IER, Euler, and Heun steps on generation quality using CelebA-HQ (\(256\times 256\) px) and ImageNet (\(256\times 256\) px). We quantitatively measure sample quality with the Fr\'echet Inception Distance (FID)~\citep{10.5555/3295222.3295408,Seitzer2020FID} and Inception Score (IS)~\citep{salimans_improved_2016}, and the diversity using the recall score~\citep{kynkaanniemi_improved_2019}. The FID for ImageNet is computed using~\citep{dhariwal2021diffusion} and the FID for CelebA-HQ (\(256\times 256\) px) and CIFAR-10 are computed by the PyTorch Fidelity library~\citep{obukhov2020torchfidelity} using 50,000 generated samples. 
\begin{figure}[t]
  \centering
  \begin{subfigure}[t]{0.99\linewidth}
    \centering
    \includegraphics[width=\linewidth]{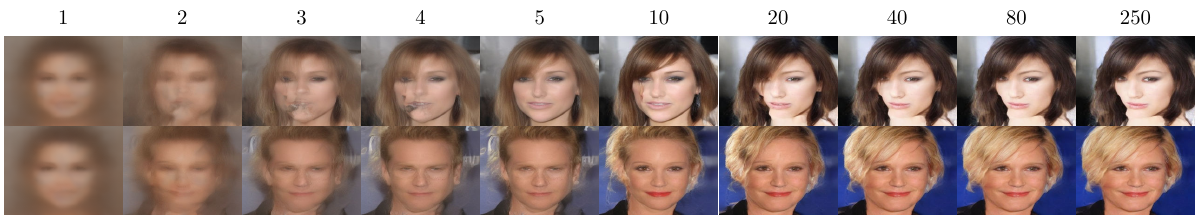}
    \caption{Effect of Euler steps on sample quality.}
    \label{fig:gen_progress_celeba}
  \end{subfigure}\hfill
  \begin{subfigure}[t]{0.99\linewidth}
    \centering
    \includegraphics[width=\linewidth]{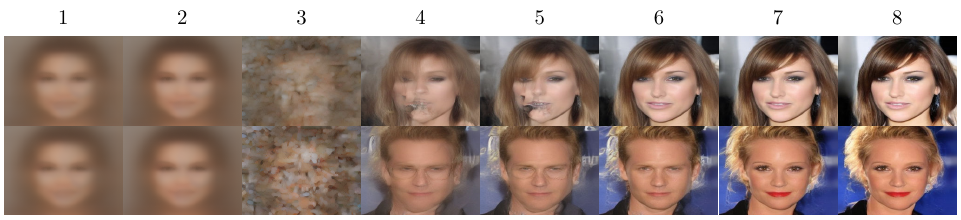}
    \caption{Effect of IER steps on sample quality.}
    \label{fig:gen_progress_celebb}
  \end{subfigure}%
  \caption{Effect of the number of forward steps on sample quality using CelebA-HQ (\(256\times 256\) px).}
\end{figure}%

As shown in Figure~\ref{fig:gen_progress_celeba}, with very few Euler steps (\(1-5\)) the samples are blurry but roughly coherent; increasing the steps to \(5-40\) significantly improves the quality and semantic detail. The FID (calculated using Heun, Figure~\ref{fig:fid_chart_celeb_imagenet}a) value decreases rapidly up to about \(40\) steps. Beyond \(40\) steps, the curve exhibits clear diminishing returns: FID improves from about \(7.45\) at \(40\) steps to only \(7.27\) at \(80\) steps. 
In contrast, the IER sampler, of which samples are presented in Figure~\ref{fig:gen_progress_celebb}, converges with far fewer steps: requiring only \(6-8\) steps for high-quality generation; and additional IER steps provide diminishing results.
\begin{table}[!t]
    \centering
    \tiny
    \caption{FID results on ImageNet \(256\times256\), both without and with classifier-free guidance (CFG). IS, precision, and recall are reported for the CFG settings.}
    \label{tab:fid_imagenet}
    \begin{tabular}[t]{lccccccc}
        \toprule
        Method & Params(M) & Training Steps & FID & FID(CFG)\(\downarrow\) & IS\(\uparrow\) & Prec.\(\uparrow\) & Rec.\(\uparrow\) \\
        \midrule
        DiT-XL/2 & 675 & 7M & 9.6 & 2.27 & 278.24 & 0.67 & \textbf{0.67} \\
        SiT-XL/2 & 675 & 7M & 8.30 & \textbf{2.06} & {277.50} & 0.68 & \textbf{0.67} \\
        \textbf{GAF-XL (ours)} & 694 & 7M & 7.85 & 2.25 & 338.00 & \textbf{0.77} & 0.63 \\
        \textbf{GAF-XL (retrunked, ours)} & 694 & 150k & \textbf{7.51} & 2.63 & 338.00 & \textbf{0.77} & 0.62 \\
        \bottomrule
    \end{tabular}
\end{table}
\begin{table}[!t]
    \begin{minipage}[t]{0.4\linewidth}
        \vspace{0pt}
        \centering
        \captionsetup{type=table}
        \tiny
        \setlength{\tabcolsep}{3pt}
        \caption{FID comparison on CelebA-HQ \(256\times256\) and CIFAR-10.}
        \label{tab:fid_celeb_and_cifar}
        \begin{minipage}[t]{0.55\linewidth}
            \centering
            \begin{tabular}{@{}lc@{}}
                \toprule
                \multicolumn{2}{c}{CelebA-HQ 256} \\
                \midrule
                Method & FID\(\downarrow\) \\
                \midrule
                Glow~\citep{kingma_glow_2018} & 68.93 \\
                SDE~\citep{song2021scorebased} & 7.23 \\
                LSGM~\citep{vahdat_score-based_2021} & 7.22 \\
                LDM-4~\citep{Rombach_2022_CVPR} & \textbf{5.11} \\
                \textbf{GAF} (\textbf{ours}) & 7.27 \\
                \bottomrule
            \end{tabular}
        \end{minipage}\hfill
        \begin{minipage}[t]{0.45\linewidth}
            \centering
            \begin{tabular}{@{}lc@{}}
                \toprule
                \multicolumn{2}{c}{CIFAR-10} \\
                \midrule
                Method & FID\(\downarrow\) \\
                \midrule
                DDPM~\citep{ho_denoising_2020} & 3.17 \\
                SDE~\citep{song2021scorebased} & \textbf{2.20} \\
                Flow Match.~\citep{lipman_flow_2023} & 6.35 \\
                \textbf{GAF} (\textbf{ours}) & 9.53 \\
                \bottomrule
            \end{tabular}
        \end{minipage}
    \end{minipage}\hfill
    \begin{minipage}[t]{0.6\linewidth}
        \vspace{0pt}
        \centering
        \includegraphics[width=\linewidth]{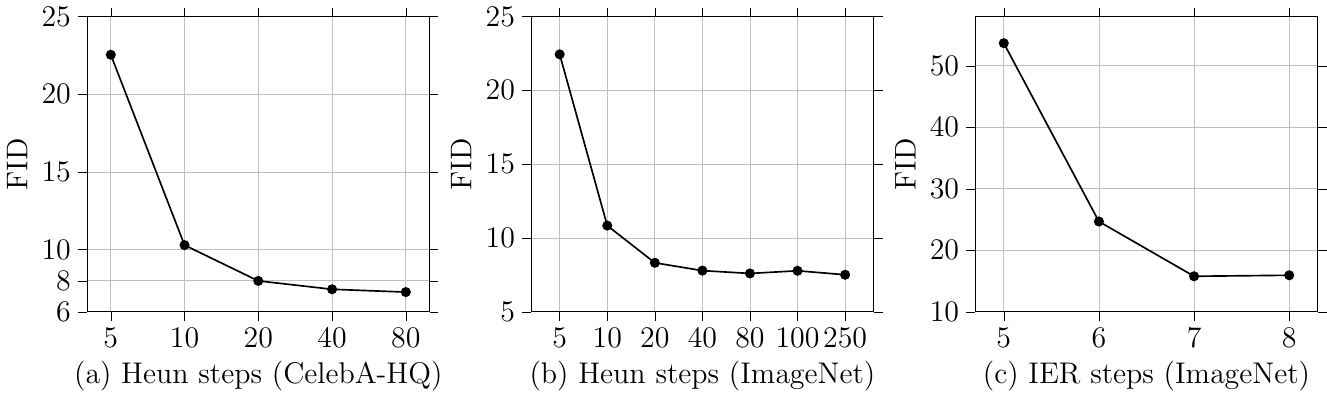}
        \captionsetup{type=figure}
        \caption{Effect of the number of IER and Heun steps on sample quality, measured by FID.}
        \label{fig:fid_chart_celeb_imagenet}
    \end{minipage}
\end{table}

Table~\ref{tab:fid_imagenet} and~\ref{tab:fid_celeb_and_cifar} report FID scores on ImageNet \(256\), CIFAR-10, and CelebA-HQ \(256\) using the Heun sampler. On ImageNet \(256\), GAF achieves an FID of \(7.85\) (at \(N=55\) steps), outperforming DiT-XL/2 (\(9.6\)) and SiT-XL (\(8.3\)), with competitive IS, precision and recall.  On CelebA-HQ \(256\), GAF achieves an FID of \(7.27\) (at \(N=80\) steps), trailing LDM-4 (\(5.11\)) which employs heavily optimized diffusion pipeline. GAF achieves an FID of \(9.53\) on CIFAR-10 dataset without CFG.

\textbf{Retrunking.} As an additional ImageNet multi-class experiment, we initialize GAF's trunk from pretrained DiT-XL/2 weights and train the trunk and the \(J/K\) heads jointly. We refer to this as \textit{retrunking}: repurposing an existing pretrained backbone as GAF's shared trunk. The trunk and endpoint heads are architecturally independent, so any compatible pretrained backbone can be adopted as the starting point for GAF training. Retrunked from DiT-XL/2, GAF achieves FID 2.63 (CFG=1.25) on ImageNet 256\(\times\)256 trained for 150,000 iterations.

\section{Transport Algebra}
\label{sec:transport_algebra}
\begin{figure}[!htbp]
  \centering
  \includegraphics[width=\linewidth]{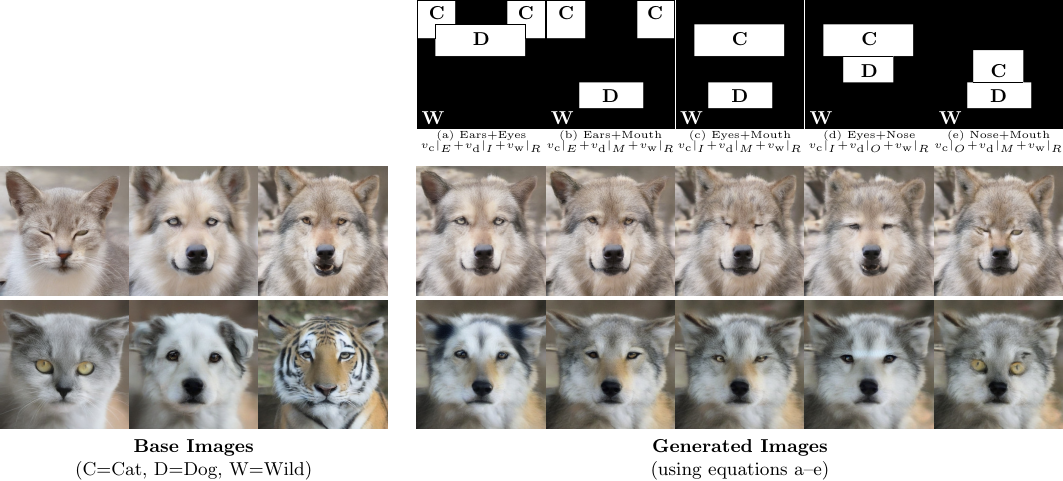}
  \caption{Spatial composition on AFHQ using masked velocity blending.}
  \label{fig:afhq_spatial_velocity_interpolation}
\end{figure}

\begin{figure}[!htbp]
  \centering
  \includegraphics[width=\linewidth]{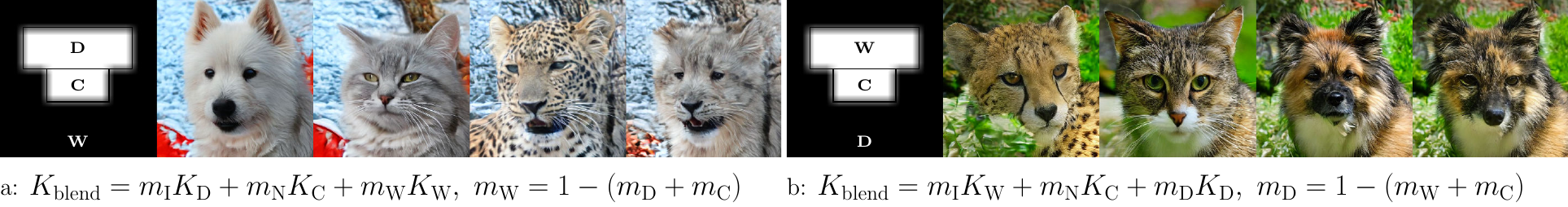}
  \caption{Spatial composition on AFHQ using masked \(K\) head blending (8-step IER).}
  \label{fig:afhq_ier_spatial_velocity_interpolation}
\end{figure}

\begin{figure}[!htbp]
  \centering
  \includegraphics[width=\linewidth]{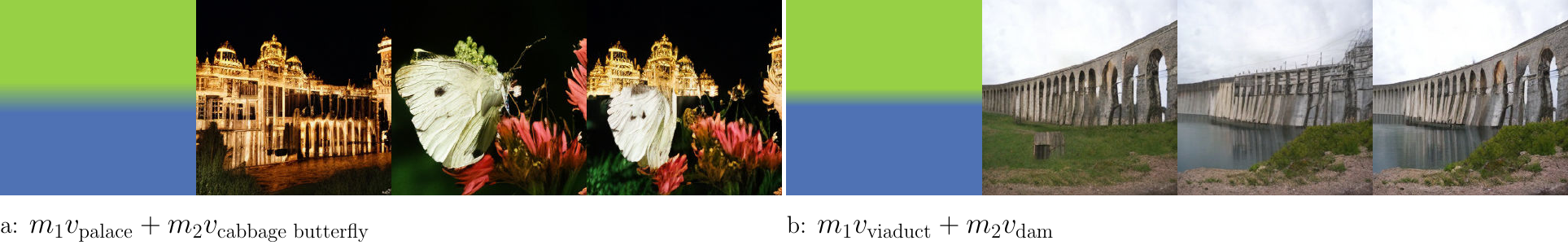}
  \caption{Spatial composition on ImageNet using masked velocity blending.}
  \label{fig:imgnet_spatial_velocity_interpolation}
\end{figure}

\paragraph{Spatial Endpoint and Velocity Interpolation.}\label{GAF:spatial_velocity_interpolation}
Figure~\ref{fig:afhq_spatial_velocity_interpolation},~\ref{fig:afhq_ier_spatial_velocity_interpolation}, and~\ref{fig:imgnet_spatial_velocity_interpolation} show that spatial composition can be performed either in endpoint space using \(K\) heads or in velocity space using \(v\). Using region masks (\(m\)), we combine class-specific endpoint heads directly (e.g., (\(K_{\text{C}}, K_{\text{D}}, K_{\text{W}}\), for Cat, Dog, and Wild classes, respectively)) to assign different semantics to different facial regions, while keeping the remaining region fixed. For example, we can assign masks to target specific region by dividing the area into \(N\in\mathbb{R}\) segments horizontally, diagonally, radially, or in any other configuration, as shown in Figure~\ref{fig:imgnet_spatial_velocity_interpolation}. We can also assign masks to regions such as ears (E), eyes (I), mouth (M), nose (O), and rest (R) (the remaining region not covered by the masks), as shown in Figure~\ref{fig:afhq_spatial_velocity_interpolation}, to composite features that are directed by the IER or velocity space integration. This produces coherent localized edits that preserve global identity in non-targeted regions, indicating that endpoint composition is a native control mechanism in GAF. Endpoint mixing offers a direct coordinate-level generation (see Figure~\ref{fig:afhq_ier_spatial_velocity_interpolation}), while velocity mixing uses the ODE generation (see Figure~\ref{fig:afhq_spatial_velocity_interpolation} and~\ref{fig:imgnet_spatial_velocity_interpolation}). Additional results are shown in Appendix~\ref{sec:spatial_mixing}.

\section{Conclusion}
We introduced Generative Anchored Fields (GAF), an endpoint prediction generative model. By learning endpoints \(J\) and \(K\) and deriving an emergent velocity field as \(v=K-J\), GAF supports flexible inference methods within a single framework: endpoint iterative refinement (IER), ODE integration, and hybrid switching between them. Since GAF exposes its endpoints directly, generation can be treated as algebra over learned transport coordinates. We demonstrated that GAF achieves FID \(2.25\) on ImageNet \(256\times 256\) with classifier-free guidance. GAF's factorization into a trunk, \(J\), and \(K\) yields inherent modularity. Since GAF separates representation learning (trunk) and transport parametrization (\(J/K\) heads), pretrained vision backbones can serve as GAF's shared trunk, inheriting Transport Algebra through \(J/K\) head attachment without retraining from scratch, a  process we call \textit{retrunking}.

\bibliographystyle{plainnat} %
\bibliography{references.bib}


\appendix



\section{Emergent Transport Dynamics}
\label{emergent_transport_dynamics}
From~\eqref{eq:velocity}, we have \(v(\mathbf{x}_t, t) = (2t-1)\mathbf{x}_t + (K_{\text{res}}-J_{\text{res}})\). Let \(\Delta_{\text{res}} = K_{\text{res}}-J_{\text{res}}\) and examine \(v\) at \(t\in[0, 1]\).
\begin{align}\label{eq:velocity_at_each_time}
    v(\mathbf{x}_t,t) = 
    \begin{cases}
        -\mathbf{z}_y+\Delta_{\text{res}} & \text{as } {t \to 0,}\\
        \Delta_{\text{res}} & t=1/2,\\
        \mathbf{z}_x+\Delta_{\text{res}} &\text{as } {t \to 1}
    \end{cases}
\end{align}

Similarly, from the swap operator in~\eqref{eq:swap_function}, we have \(v(\mathbf{x}_{1-t},t)=(1-2t)\mathbf{x}_{1-t} + (\tilde{K}_{\text{res}}-\tilde{J}_{\text{res}})\). Let \(\tilde{\Delta}_{\text{res}}=\tilde{K}_{\text{res}}-\tilde{J}_{\text{res}}\) and examine \(v\) at \(t \in [0,1]\).
\begin{align}\label{eq:swap_velocity_at_each_time}
v(\mathbf{x}_{1-t},t) = 
\begin{cases}
\mathbf{z}_y+\tilde{\Delta}_{\text{res}} & \text{as } {t \to 0,}\\
\tilde{\Delta}_{\text{res}} & t=1/2,\\
-\mathbf{z}_x+\tilde{\Delta}_{\text{res}} &\text{as } {t \to 1, }
\end{cases}
\end{align}

\section{Transport Algebra: Pairing Topologies}
\label{J_K_topologies}
\begin{figure}[!htbp]
    \centering
    \includegraphics[width=0.85\linewidth]{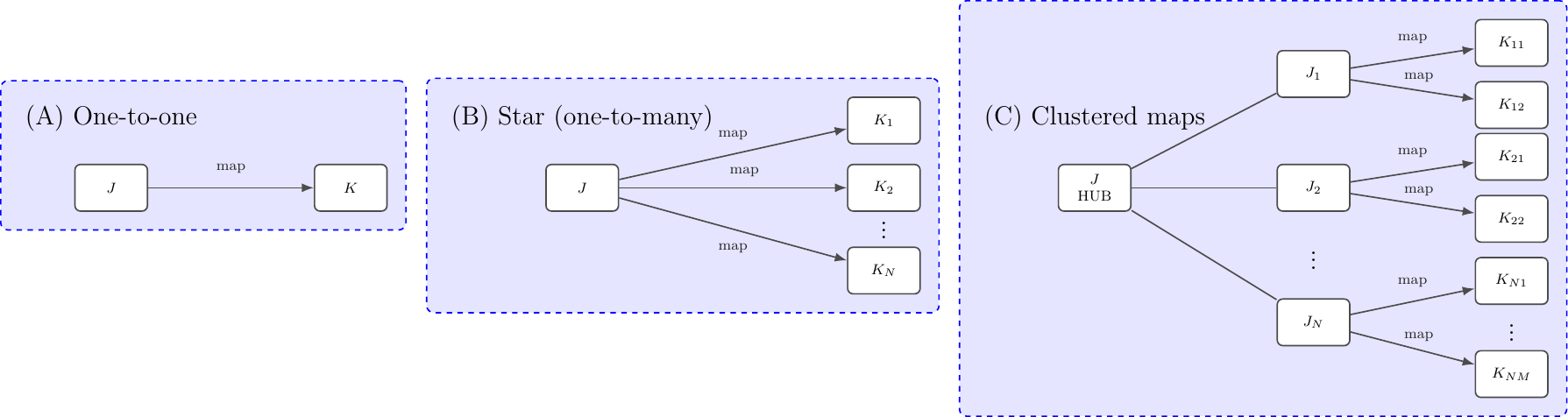}
    \caption{Three \(J/K\) pairing topologies: (A) one-to-one, (B) star (one-to-many), and (C) clustered. Each topology shows different ways to map the noise endpoint \(J\) to a data endpoint \(K\).}
    \label{fig:noise_map}
\end{figure}

\paragraph{\(J/K\) pairing topologies.} The swap operator \(\mathcal{S}\) gives us the mechanism to traverse through the endpoints back and forth in time. In addition, we can also consider the three main different ways multiple pairs of twins \(J\) and \(K\) can be mapped into \(\{(J_{n},K_{n})\}_{n=1}^N\) mappings. The pairings are: (A) one-to-one, (B) star (one-to-many), and (C) clustered mapping, shown in Figure \ref{fig:noise_map}. 
\label{GAF:NetwArch}

\paragraph{Cross-modality transport.} 
\label{cross_modality_transport}
The \(v_{i\to j}\) expands to an encode-decode process: 
     
    \begin{enumerate}
        \item Decode to class \(i\): Integrate forward with \(v_{J\to i} = K_i - J\) to reach \(K_i\) (forward time)
        \item Encode to noise: Integrate with \(-v_{i\to J} = J - K_i\) to reach \(J\) (reverse time)
        \item Decode to target class \(j\): Integrate forward with \(v_{J\to j} = K_j - J\) to reach \(K_j\) (forward time)
        
        The full transport path is:   
        \begin{equation}\notag
            v_{i\to j}=K_i \xrightarrow{-(K_i-J)} J \xrightarrow{K_j-J} K_j
        \end{equation}
    \end{enumerate}
This sequential transport process uses \(J\) as a common intermediate representation shared across all classes, enabling transport and composition between any pair of \(K\) heads.

\section{Sampling}
\subsection{ODE Sampling}\label{GAF:Sampling-ODE}
Recall from Section~\ref{GAF:Sampling} that we can use the novel IER sampler for sampling from GAF. We can also sample from GAF using an ODE. At inference, we generate samples by integrating the emergent velocity field \(v=K-J\) from noise (\(t=0\)) to data (\(t=1\)) using the standard Euler's method~\citep{10.5555/3600270.3602196}. Unlike standard models, GAF defines this field algebraically, enabling precise compositional generation directly within the ODE solver. We discretize the time interval [\(t_\text{eps}, 1-t_\text{eps}\)] into uniformly spaced steps: 
\begin{equation}
    t_0=t_\text{eps},t_1,\dots,1-t_\text{eps}\notag, \text{with} ~t_\text{eps}=1\times10^{-3}
\end{equation}

At each step, we update the latent state via
\begin{equation}
    z_{k+1}=z_k + (t_{k+1} - t_k)v(z_k,t_k).
\end{equation}

We initialize \(\mathbf{z}_0\sim \mathcal{N}(\mathbf{0},\mathbf{I})\) at \(t=t_\text{eps}\) and evolve to \(t=1-t_\text{eps}\). For multi-class generation, samples are conditioned on their target class label \(y\) through the corresponding \(K_n\) heads.

\subsection{Hybrid Sampling}\label{GAF:Sampling-Hybrid}
GAF supports mid-trajectory switching between IER and ODE at any step, in both forward (generation) and reverse (inversion) directions. Run Algorithm~\ref{alg:ier_forward} until \(t_s\), then continue with Algorithm~\ref{alg:ier_reverse}, or vice versa.

\section{Implementation Details}
\label{sec:implementation_details}
\subsection{Network Architecture}
\label{detailed_arch_desc} 
\begin{figure}[!htbp]
  \centering
  \includegraphics[width=0.99\linewidth]{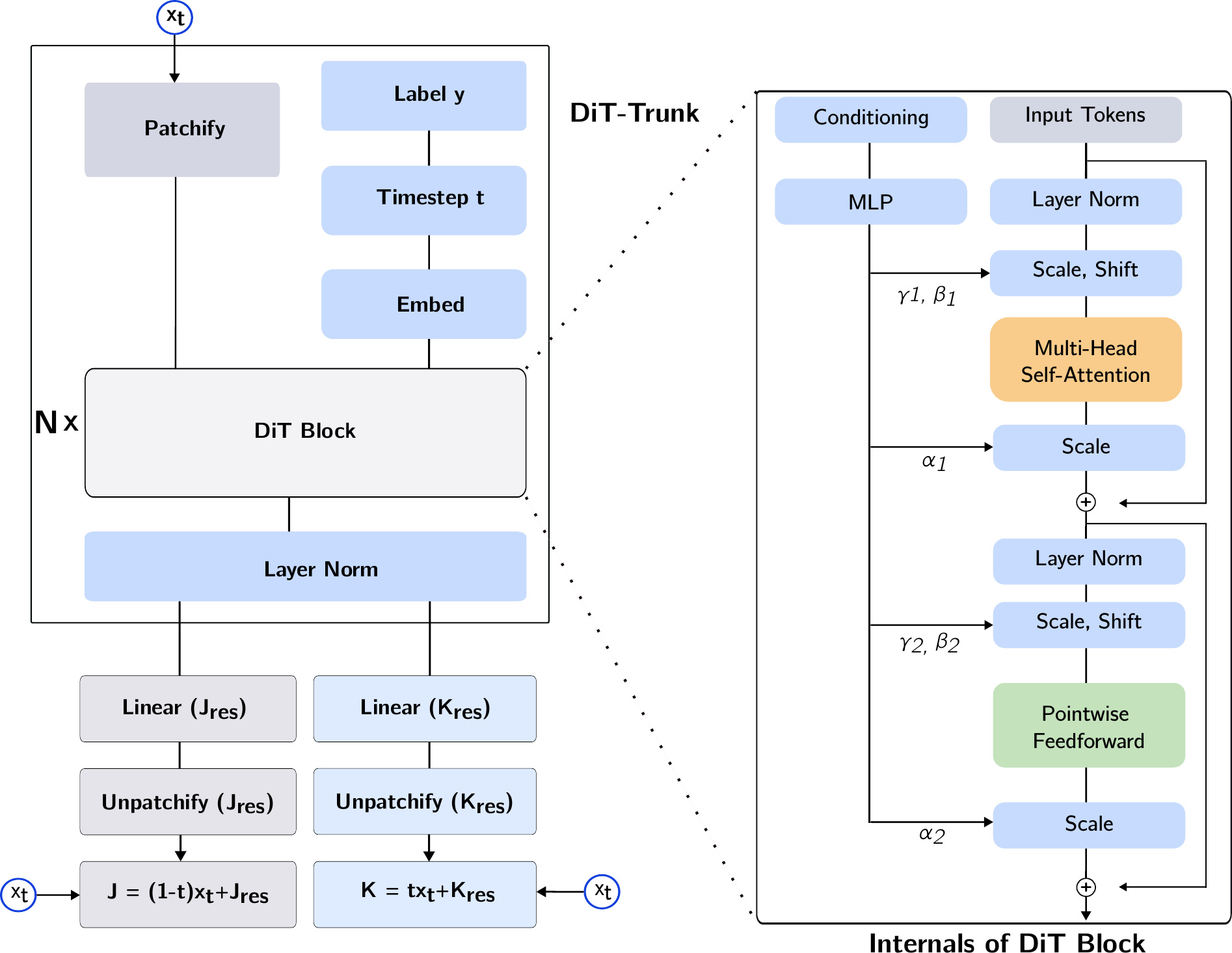}
  \caption{GAF architecture for image generation. The DiT-Trunk block follows the architecture of~\citep{williampeebles2023}. GAF replaces DiT's single output projection with two independent heads, \(H_J\) and \(H_K\) (see Section~\ref{GAF:twins}), each producing endpoint residuals that are unpatchified separately and combined with scaled input to form the twins \(J\) and \(K\).}
  \label{fig:GAF_main_architecture}
\end{figure}

An overview of the architecture for image generation is given in Figure~\ref{fig:GAF_main_architecture}. We discuss each component below.

\paragraph{Trunk.} As shown in Figure~\ref{fig:GAF_main_architecture}, we employ a Diffusion Transformer (DiT)~\citep{williampeebles2023} architecture as our feature extractor. The GAF trunk takes in either a pixel-space representation (e.g., \(32\times32\times3\) px for low-resolution images) or a latent-space representation (e.g., \(32\times32\times4\) latents, encoded from high-resolution images such as \(256\times 256\times 3\) px using a pretrained variational autoencoder (VAE)~\citep{kingma_auto-encoding_2013}). That is because, at higher resolution, pixel-space processing becomes impractical due to memory and compute constraints.

\paragraph{Patchify.} Patchify is the first layer of the trunk, converting the pixel- or latent-space inputs into a sequence of \(T\) tokens. It does this by dividing the input in patches, and linearly embedding each patch into the trunk's hidden dimension \(D\). Standard sinusoidal positional embeddings are then added to the tokens to retain spatial information. As noted in~\citep{williampeebles2023}, the choice of patch size \(p\) presents a crucial tradeoff between computational cost and model granularity, as halving \(p\) quadruples the GFLOPS with negligible effect on model parameter count. Therefore, we set \(p\) as \(2\) in our configurations.

\paragraph{DiT Block.} The trunk processes the patch-embedded inputs through \(N\) transformer blocks, each consisting of multi-head self-attention and feed-forward layers with adaptive layer normalization (adaLN). Conditioning on timestep \(t\) and class label \(y\) is achieved by embedding each into \(D\)-dimensional vectors. Timestep \(t\) is mapped through a sinusoidal encoding followed by a two-layer MLP. These embeddings are summed to form the conditioning vector \(c = t_{emb} + y_{emb}\), which modulates each transformer block via adaLN.
Following \(N\) transformer blocks, a final adaLN layer conditioned on \(c\) produces feature tokens \(\mathbf{f}_t\in{\mathbb{R}^{N \times D}}\).

\paragraph{J Head (Noise Boundary.)} The \(J\) head is a single-layer \(\text{MLP}_{J}\), which maps trunk features to noise boundary residuals. The MLP expands from \(D\) to \(2D\) hidden units via a linear layer, and then projects to \(p^2\times C\) features, where \(C\) is the number of channels (\(4\) for latents, 3 for RGB pixels). The final layer of \(\text{MLP}_J\) is initialized to zero for stability. Finally, the outputs are reshaped (i.e., unpatchified) back to their spatial form (\(C\times H\times W\)). 

\paragraph{K Heads (Data Boundaries).} For each class \(n \in \{0,\dots,N-1\}\), each corresponding \(K\) head is an independent single-layer \(\text{MLP}_K\). The \(K\) heads share no parameters across classes, enabling independent transport operators for each class. During the forward pass, trunk features are computed once for the entire batch, then routed to the appropriate \(K\) heads based on class labels. The routing ensures each \(K\) head trains on only its assigned class, while the trunk learns shared representations across all classes. Therefore, a single linear projection is sufficient for high-fidelity generation, confirming that the trunk \(\Phi\) captures the necessary feature information for the twins.


\subsection{Experimental Setup}
\label{sec:evaluation-setup}
\subsubsection{Dataset} We train and evaluate GAF on four datasets. 
\label{sec:training_data}

\textbf{CIFAR-10}~\citep{krizhevsky_learning_nodate} contains 50,000 training images across 10 object classes at \(32\times 32\) px 
resolution, with \(5,000\) images per class. We use this dataset for a \(10\)-class (i.e., one \(J\) and ten \(K\)) generation task.

\textbf{CelebA-HQ}~\citep{liu2015faceattributes} consists of \(202,599\) celebrity face images resized to \(256\times 256\) px resolutions. We use this dataset for a single-class (i.e., one \(J\) and one \(K\)) generation task.

\textbf{AFHQ}~\citep{choi2020starganv2} contains \(15,000\) high-quality images of cats, dogs, and wildlife at \(512\times 512\) px resolution. We use this dataset for a \(3\)-class (i.e., one \(J\) and three \(K\)) generation task.

\textbf{ImageNet-1k}~\citep{deng2009imagenet,ILSVRC15} is the full ImageNet classification dataset with \(1.28\) million training images across \(1,000\) object classes, with \(1,300\) images per classes. For demonstration purposes, we use a single-class dataset (\textit{school bus}). Additionally, we use a multi-class subset consisting of \(1,000\) distinct classes, for which we use \(1,000\) independent \(K\) heads, each corresponding to a specific class.

\subsubsection{Training}
\label{sec:training_setup} 

\label{training_loss_desc} 
We train GAF with three loss components defined in~\eqref{eq:GAF_loss} using mixed-precision (bfloat16) training.
\begin{equation}
\mathcal{L}_{\text{GAF}}
= \mathcal{L}_{\text{pair}}
+ \lambda_{\text{res}}\,\mathcal{L}_{\text{res}}
+ \lambda_{\text{swap}}\,\mathcal{L}_{\text{swap}}\notag,
\label{eq:GAF_loss_ref}
\end{equation}

where \(\mathcal{L}_{\text{pair}}\) anchors the twins \(J\) and \(K\) to their endpoints (\(\mathbf{z}_y\) and \(\mathbf{z}_x\)), \(\mathcal{L}_{\text{res}}\) penalizes the residuals near endpoints (as \(t\rightarrow0\) for \(J\) and as \(t\rightarrow1\) for \(K\)), and \(\mathcal{L}_{\text{swap}}\) enforces time-reversal symmetry at the midpoint for consistency and self-correction. 

For each training sample, we draw a uniform timestep \(t\) from \(\text{Uniform}[t_{\text{eps}},1-t_{\text{eps}}]\), with \(t_{\text{eps}}=1\times10^{-3}\). We compute the forward prediction at \(t\) for the main loss term, and evaluate it once more at the complementary timestep \(1-t\); this time-reversed prediction at \(1-t\) is used to compute the swap loss.

The complete training configurations for each dataset are shown in Table~\ref{tab:training_config}.
Through an extensive parameter sweep of the residual and swap loss coefficients, we found \(\lambda_{\text{res}}=0.003\) and \(\lambda_{\text{swap}}=0.002\) to perform best. We use these values in all experiments unless stated otherwise. The training images are first encoded by the SD1.5-VAE~\citep{Rombach_2022_CVPR}. We use the AdamW optimizer with \(\beta=(0.9,0.99)\), and learning rate \(1\times10^{-4}\) for all of the datasets, with no weight decay in any of the training runs. The ImageNet-1k school bus class is trained for \(300,000\) iterations, whereas the other models are trained for \(100,000\) iterations.
\begin{table}[!htbp]
    \centering
    \footnotesize
    \caption{Training configurations for each dataset.}
    \label{tab:training_config}
    \begin{tabular}{lccccc}
        \toprule
        & CIFAR-10 & CelebA-HQ & AFHQ & ImageNet & ImageNet \\
        \midrule
        Training & scratch & scratch & scratch & retrunked & scratch\\
        Architecture & DiT-B/2 & DiT-XL/2 & DiT-XL/2 & DiT-XL/2 & DiT-XL/2\\
        VAE & -- & SD 1.5 & SD 1.5 & SD 1.5 & SD 1.5 \\
        Batch size & 256 & 32 & 32 & 128 & 256 \\
        Iterations & 100k & 100k & 100k & 150k & 7M\\
        Learning rate & 2e-4 & 1e-4 & 1e-4 & 1e-4 & 1e-4\\
        Warmup steps & 500 & 3000 & 3000& 0 & 0\\
        \(\beta_1, \beta_2\) & 0.9, 0.99 & 0.9, 0.99 & 0.9, 0.99 & 0.9, 0.99 & 0.9, 0.99 \\
        EMA decay & 0.9995 & 0.9999 & 0.9999 & 0.9999 & 0.9999 \\
        \(\lambda_{\text{res}}\) & 0.003 & 0.003 & 0.003 & 0.0001 & 0.0001 \\
        \(\lambda_{\text{swap}}\) & 0.002 & 0.002 & 0.002 & 0.0001 & 0.0001 \\
        \bottomrule
    \end{tabular}
\end{table}

\section{Additional Results}
\subsection{Single-Class Generation (One \texorpdfstring{\(J\)}{J} and One \texorpdfstring{\(K\)}{K})}
\label{sec:evaluation-single}
\begin{figure}[!htbp]
  \centering
  \includegraphics[width=0.99\linewidth]{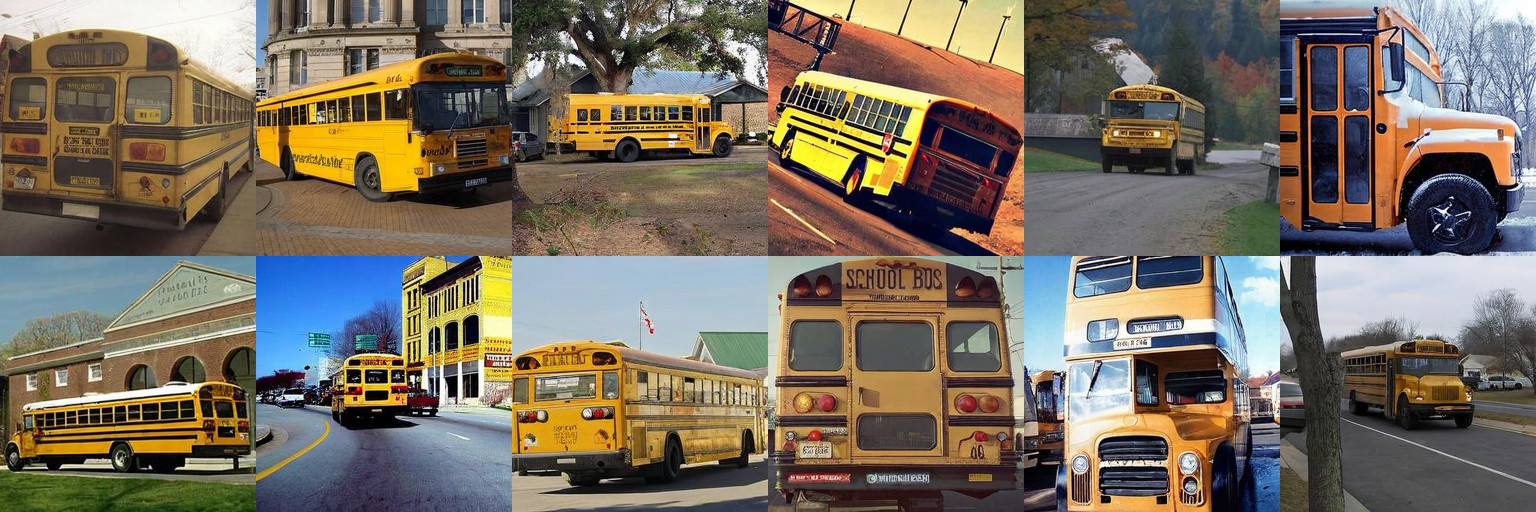}
  \caption{Single-class generated samples (\(256\times256\) px) from GAF trained on the ImageNet \textit{school bus} class.}
  \label{fig:school_bus_samples}
\end{figure}

This section validates GAF on single-class generation, using the \textit{school bus} class from ImageNet, which contains about 1.3k images used for training. Figure~\ref{fig:school_bus_samples} shows some example generations. We can see that a single-class GAF model with one \(J\) and one \(K\) head, using 250-step Heun sampling, demonstrates high-quality, diverse generations. Note that the generations showcase varied viewpoints, lighting conditions, and backgrounds while maintaining class consistency.

\subsection{Transport Algebra}
\label{sec:evaluation-transport}

This section showcases three applications of GAF's novel transport algebra: pairwise interpolation (in Section~\ref{sec:evaluation-transport-interpolation}), continuous cyclic transport (in Section~\ref{sec:evaluation-transport-cyclic}), and barycentric transport (in Section~\ref{sec:evaluation-transport-barycentric}).

\subsubsection{Pairwise Interpolation}
\label{sec:evaluation-transport-interpolation}

\begin{figure}[!htbp]
\centering
\begin{tikzpicture}
\node[anchor=south west,inner sep=0] (img) at (0,0)
  {\includegraphics[width=0.8\linewidth]{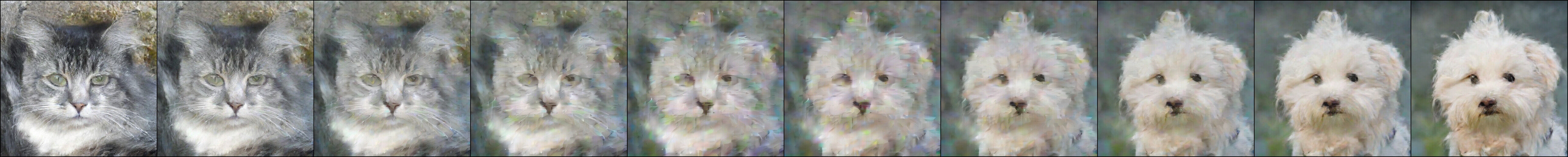}};
\begin{scope}[x={(img.south east)},y={(img.north west)}]
  \node[anchor=east] (labA) at (-0.14,0.5) {a)};
  \node[anchor=west] at ([xshift=0.3em]labA.east)
    {\(v_{\text{cat}\to \text{dog}}\)};
\end{scope}
\end{tikzpicture}
\vspace{0.1cm}
\begin{tikzpicture}
\node[anchor=south west,inner sep=0] (img) at (0,0)
  {\includegraphics[width=0.8\linewidth]{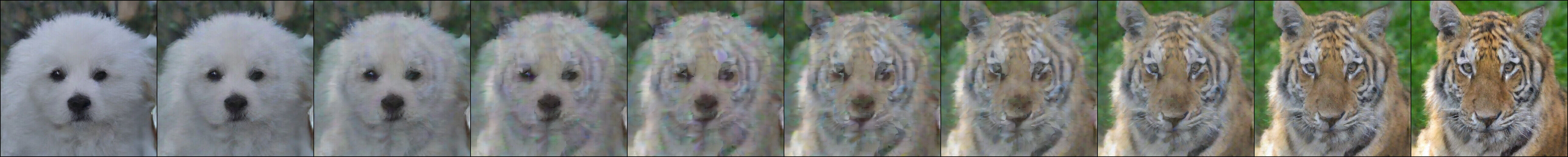}};
\begin{scope}[x={(img.south east)},y={(img.north west)}]
  \node[anchor=east] (labB) at (-0.14,0.5) {b)};
  \node[anchor=west] at ([xshift=0.3em]labB.east)
    {\(v_{\text{dog}\to \text{wild}}\)};
\end{scope}
\end{tikzpicture}
\vspace{0.1cm}
\begin{tikzpicture}
\node[anchor=south west,inner sep=0] (img) at (0,0)
  {\includegraphics[width=0.8\linewidth]{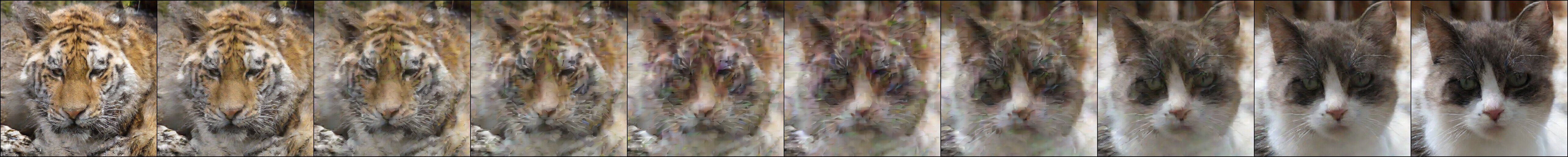}};
\begin{scope}[x={(img.south east)},y={(img.north west)}]
  \node[anchor=east] (labC) at (-0.14,0.5) {c)};
  \node[anchor=west] at ([xshift=0.3em]labC.east)
    {\(v_{\text{wild}\to \text{cat}}\)};
\end{scope}
\end{tikzpicture}
\caption{Pairwise interpolation samples (\(512\times512\) px) from GAF trained with one \(J\) and three independent \(K\) heads on the AFHQ dataset.}
\label{fig:afhq_interpolation}
\end{figure}

We demonstrate direct \(K\) head interpolation, by performing the operation \(K_{\text{interp}}=(1-\alpha)K_i + \alpha K_j\) or \(v=(1-\alpha)v_i + \alpha v_j\), for \(\alpha \in [0,1]\). Figure~\ref{fig:afhq_interpolation} shows samples from interpolating between two classes from the AFHQ dataset, trained for multi-class generation in Section~\ref{sec:evaluation-multi}. We can see a smooth semantic transition between all three class pairs. Each interpolation produces coherent intermediate states, validating that independent \(K\) heads enable linear composition directly over the K heads or in velocity space.

\subsubsection{Continuous Cyclic Transport.}
\label{sec:evaluation-transport-cyclic}
In this section, we show that multi-class continuous generation and interpolation using \(J\) as a shared hub.

\textbf{Cyclic Endpoint Transport.}
We quantitatively analyze GAF's endpoint transport by running three RK4~\citep{chen_neural_2018, 10.5555/3600270.3602196} integration steps on the model trained on the AFHQ dataset (Cat \(\to\) Dog \(\to\) Wild \(\to\) Cat, Dog \(\to\) Wild \(\to\) Cat \(\to\) Dog, and Wild \(\to\) Cat \(\to\) Dog \(\to\) Wild) and measuring LPIPS~\citep{zhang2018perceptual}, pixel MSE (post-VAE), and latent MSE (pre-VAE) between the initial and final states.

We use the cross-modality transport operator via \(J\) defined in Section~\ref{GAF:TA}. Each cycle (e.g., Cat \(\to\) Dog, Dog \(\to\) Wild, Wild \(\to\) Cat) is implemented as the encode–decode path: \(v_i = K_i - J\), using the following method:

(A) Decode from the shared noise anchor \(z_y\) to class \(i\) with \(v_i\) (forward in time),

(B) Encode back to \(J\) with \(-v_{i}\) (reverse in time), and then,

(C) Decode to the next class \(j\) with \(v_{i\to j}\).
\begin{figure}[!htbp] 
  \centering
  \includegraphics[width=0.9\linewidth]{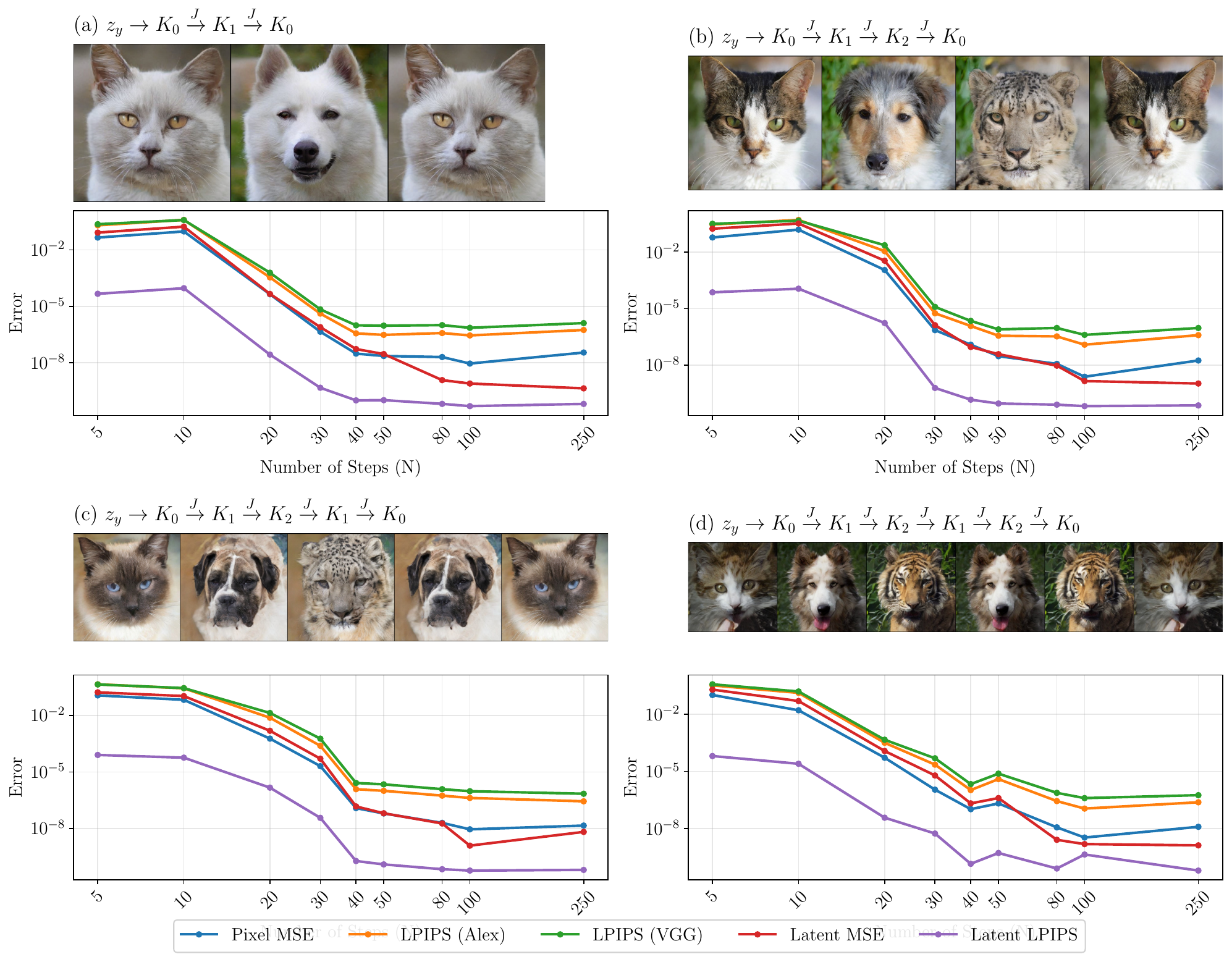}
  \caption{\textbf{Cyclic Transport Analysis.} We visualize the cyclic transport maps across classes (Top images above the error graphs) and their corresponding convergence error rates (Bottom figures showing the error graphs across multi-class generation). The trajectories correspond to cyclic transport along the cycle {\(v_{i \to j \to k \to i}\)} across 4 randomly chosen semantic cycles. The error difference between the first and the last image decreases monotonically, reaching machine precision (\(\approx10^{-16}\)) at \(N=1,000\), where \(N\) is the number of forward steps.}
  \label{fig:cycle_transport_consistency}
\end{figure}

Thus, a full Cat \(\to\) Dog \(\to\) Wild \(\to\) Cat cycle is a sequence of decode–encode legs that always returns through \(J\), as in the general cross-modality path \(v_i \xrightarrow{-(v_i)} J \xrightarrow{} v_j\). The error metrics in Figure~\ref{fig:cycle_transport_consistency} report the difference between the first initial (\(K_0\)) and last endpoint of each cycle (\(K_0\), which is the initial class itself), both in latent space (pre-VAE decoding) and image space (post-VAE decoding).

\begin{figure}[!htbp]
\centering
\begin{tikzpicture}
\node[anchor=south west,inner sep=0] (img) at (0,0) {\includegraphics[width=0.7\linewidth]{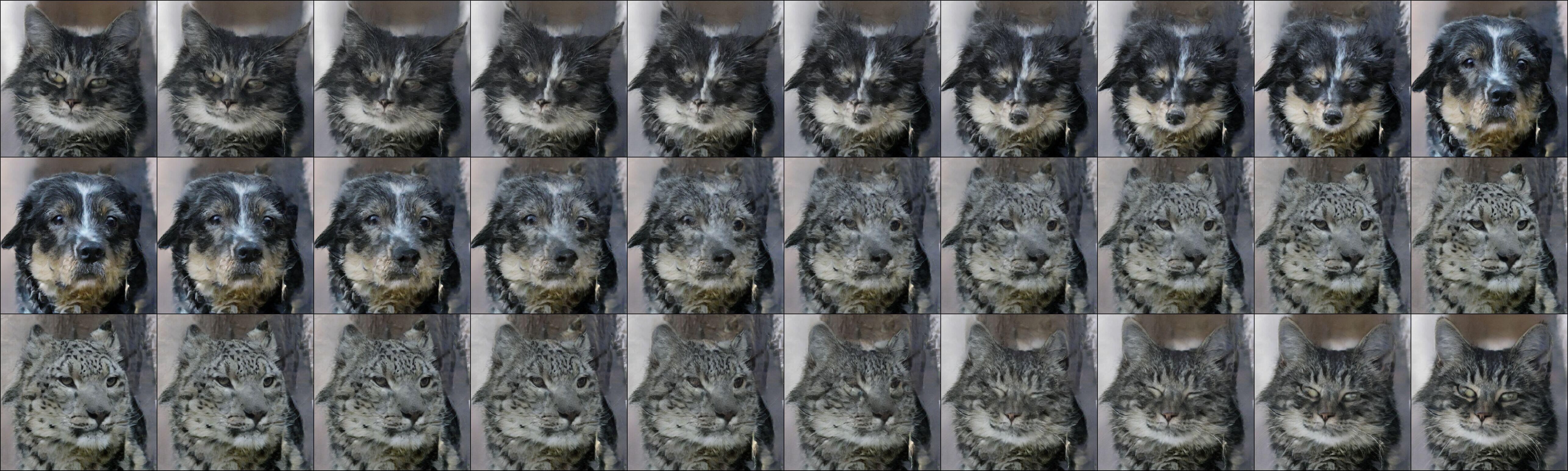}};
\begin{scope}[x={(img.south east)},y={(img.north west)}]
\node[left] at (0,5/6) {\(v_{\text{cat}_{0}\to dog}\)};
\node[left] at (0,3/6) {\(v_{\text{dog}\to wild}\)};
\node[left] at (0,1/6) {\(v_{\text{wild}\to cat_{0}}\)};
\end{scope}
\end{tikzpicture}

\caption{Continuous cyclic interpolation transport preserving identity across three classes. Each row shows a complete cycle starting from one class (left) to two interpolated classes, and finally returning to the initial class. Distinctive features (coloring, facial marking) are semantically preserved throughout each cycle while class-specific structure transforms.}
\label{fig:cyclic_transport}
\end{figure}

\textbf{Cyclic Velocity Space Interpolation.}
\label{sec:continuous_cyclic_transport}
We demonstrate that GAF enables continuous cyclic transport using \textit{cyclic velocity space interpolation}, as shown in Figure~\ref{fig:cyclic_transport}. We use the same multi-class model as in Section~\ref{sec:evaluation-multi}. For a given pair of classes (\(i, j\)), we define a parametric family of interpolated data spaces or velocity fields:
\begin{equation}\notag
    K_{i\to j} = (1-\alpha)K_i + \alpha K_j, \quad \alpha \in [0,1],
\end{equation}
\begin{equation}\notag
    v_{i\to j} = (1-\alpha)v_i + \alpha v_j, \quad \alpha \in [0,1],
\end{equation}
where \(v_i = K_i - J\) and \(v_j = K_j - J\) are the class-specific velocity fields. 
The full cyclic interpolation between the three classes \(i, j\), and \(k\) is computed as:
\begin{equation}\notag
    K_{i} \xrightarrow{J} K_{j} \xrightarrow{J} K_{k} \xrightarrow{J} K_{i}
\end{equation}
\begin{equation}\notag
    v_{i} \xrightarrow{-v_{i}} J \xrightarrow{ } v_{i\to j} \xrightarrow{-v_{i\to j}} J \xrightarrow{ } v_{j\to k} \xrightarrow{-v_{j\to k}} J \xrightarrow{ } v_{i}
\end{equation}

We sample \(\alpha\) at \(10\) uniformly spaced values in \([0,1]\) and, for each \(\alpha\), refine the endpoint or integrate the velocity field by recovering the noise endpoint from the previous data endpoint using IER or the same ODE solver. Class \(i\) is the source class and \(j\) is the target class; the interpolated velocity field \(v_{i\to j}\) moves from the \(i\)-manifold toward the \(j\)-manifold while changing only the velocity field, not the underlying integration scheme or random seed. The velocity field \(v_{i\to j}\) maintains semantic coherence throughout the transition across classes.

Across all three cycles and \(3,000\) random \(z_0\) samples, we observe an average latent LPIPS of \(\approx\) \(1.3\times 10^{-16}\) between the initial and the final images in each cycle (\(N=1,000\)). This indicates GAF's transport algebra is deterministically reversible. The near-perfect closure validates that independent \(K\) heads form a consistent algebraic structure where cyclic compositions return exactly to their starting point.

Figure~\ref{fig:cyclic_transport} demonstrates continuous identity-preserving transport through multiple classes. Starting from an initial sample from the first class, we sequentially interpolate through two additional classes before returning to the original class, using velocity blending for each pairwise transition (as in Section~\ref{sec:evaluation-transport-interpolation}). The endpoint of each interpolation becomes the starting point of the next, maintaining latent continuity throughout the cycle. 

\subsubsection{Spatial Mixing}
\begin{figure}[!htbp]
  \centering
  \includegraphics[width=0.75\linewidth]{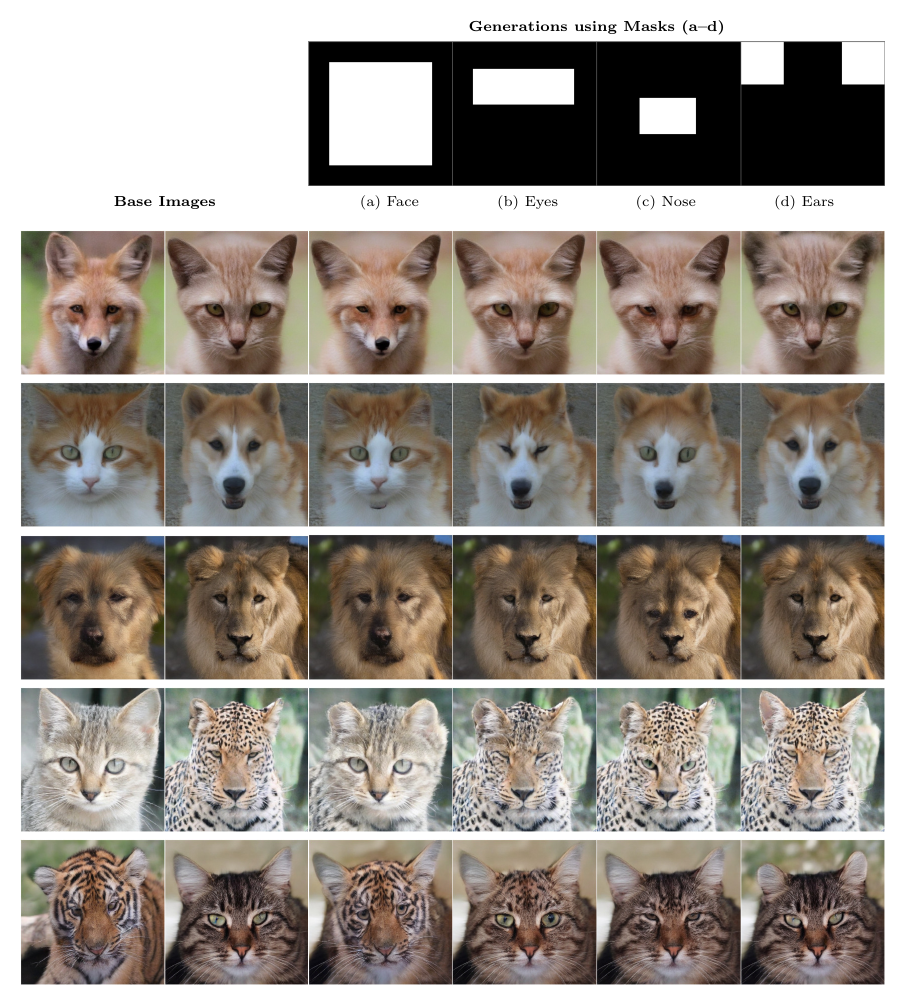}
  \caption{\textbf{Spatial mixing results.} Each column shows base images (left sub-column) and the corresponding generations (right sub-column) where structure is transferred according to one of four spatial masks: (a) face, (b) eyes, (c) nose, (d) ears.}
  \label{fig:spatial_mixing}
\end{figure}

As shown in Section~\ref{sec:transport_algebra}, we can generate novel features between classes by applying \(K\) heads to different regions using spatial masks. Figure~\ref{fig:spatial_mixing} provides additional examples on AFHQ, showing GAF enables precise feature level transfer across the face, eyes, nose, and ears regions.
\label{sec:spatial_mixing}
\subsubsection{Barycentric Transport}
\label{sec:evaluation-transport-barycentric}
We extend pairwise interpolation to a three-way composition using barycentric coordinates. The blended velocity field is \(v_{\text{blend}} = \alpha v_i + \beta v_j + \gamma v_k\), where \(\alpha + \beta + \gamma = 1\), see Section~\ref{GAF:multi_class_composition}. We use square-to-simplex mapping with \(\alpha=u(1-u), ~\beta=(1-u)(1-v),\) and \(\gamma=v\) to cover the full compositional space.
\begin{figure}[!htbp]
    \centering
    \begin{subfigure}[t]{0.3\linewidth}
        \centering
        \includegraphics[width=0.99\linewidth]{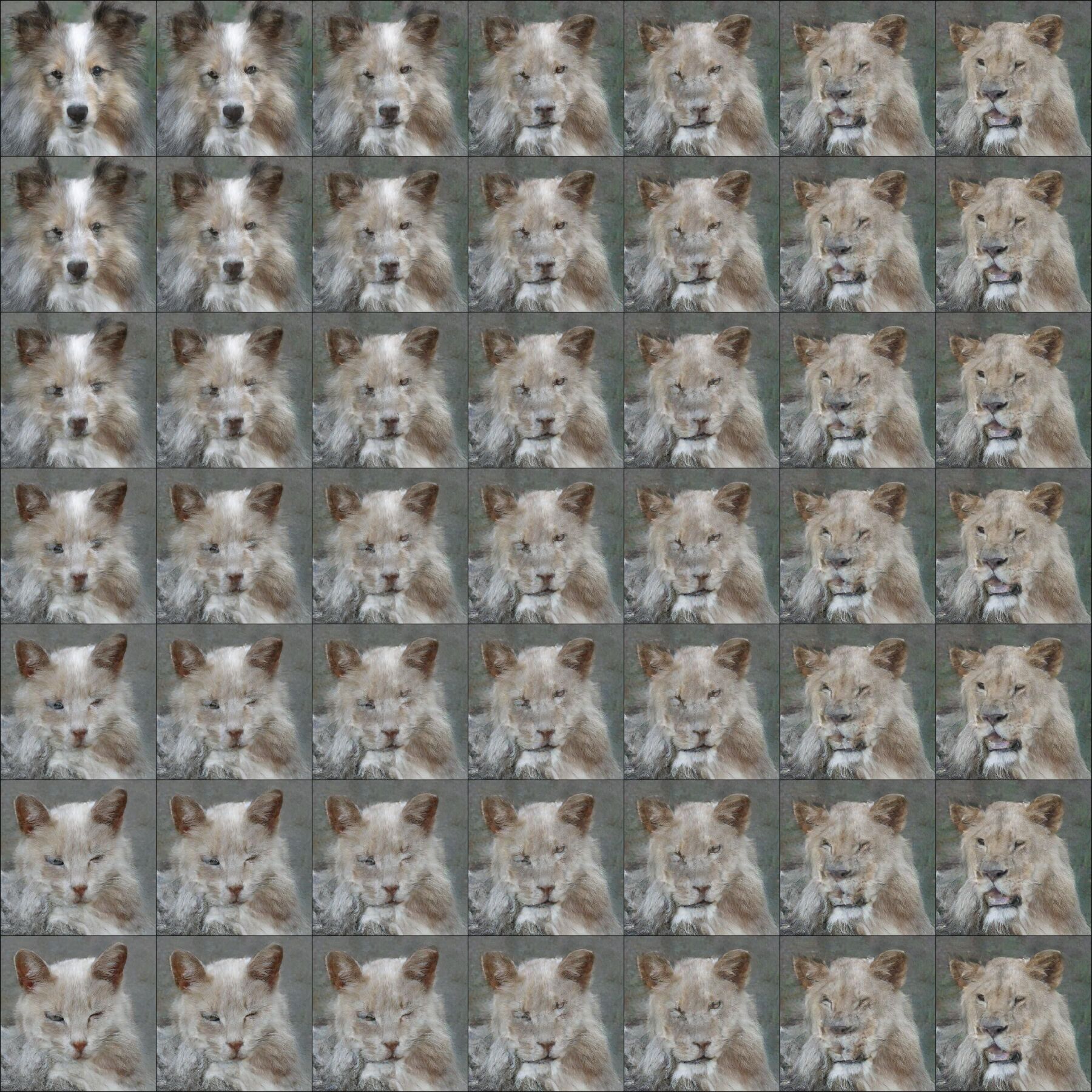}
        \caption{Dog-cat-wild.}
    \end{subfigure}%
    \hspace{0.12em}%
    \hspace{0.12em}%
    \begin{subfigure}[t]{0.3\linewidth}
        \centering
        \includegraphics[width=0.99\linewidth]{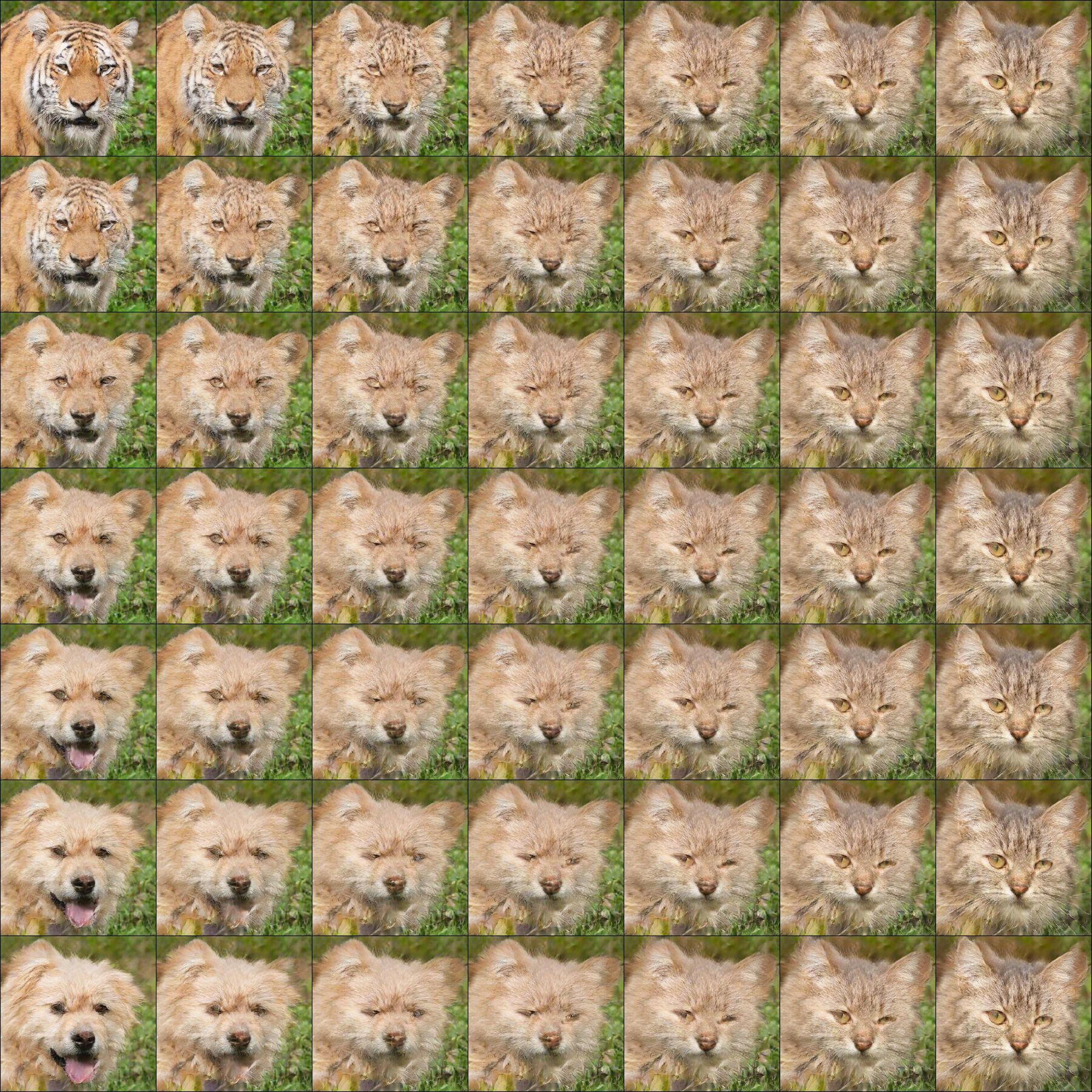}
        \caption{Wild-dog-cat.}
    \end{subfigure}%
    \vspace{0.12em}
    \begin{subfigure}[t]{0.35\linewidth}
        \centering
        \begin{tikzpicture}[scale=2.5]
            \draw[thick] (0,0) rectangle (1,1);
            \draw[step=0.166, gray!10, thin] (0,0) grid (1,1);
            \draw[->, thick, blue!40] (0, 1.2) -- (1, 1.2) node[midway, above, font=\small, text=blue!60] {\(v\)};
            \node[text=blue!60, font=\small] at (0, 1.2) [left] {0};
            \node[text=blue!60, font=\small] at (1, 1.2) [right] {1};
            \draw[->, thick, red!40] (-0.1, 1) -- (-0.1, 0) node[midway, left, font=\small, text=red!60] {\(u\)};
            \node[text=red!60, font=\small] at (-0.1, 1) [above] {0};
            \node[text=red!60, font=\small] at (-0.1, 0) [below] {1};
            \node[anchor=south west, align=left, font=\small] at (0,1) {{Class 1} (\(\beta=1\))};
            \node[anchor=north west, align=left, font=\small] at (0,0) {{Class 0} (\(\alpha=1\))};
            \draw[line width=1pt, green!50!black!30] (1,0) -- (1,1);
            \node[anchor=west, align=left, font=\small, color=black] at (1.05, 0.5) {{Class 2}\\(\(\gamma=1\))};
        \end{tikzpicture}
        \caption{Interpolation grid map showing the mixing ratios.}
    \end{subfigure}
    \caption{Barycentric Transport Algebra. Visualization of three-way velocity blending across the Dog, Cat, and Wild classes. \textbf{(a and b)} \(7\times 7\) grids showing the semantic simplex. Corners represent pure classes; interior points show linear interpolations of the vector fields. \textbf{(c)} The corresponding weight map.} 
    \label{fig:barycentric_interpolation}
\end{figure}

Figure~\ref{fig:barycentric_interpolation} shows three examples of three-way compositions, namely dog-cat-wild, cat-dog-wild, and wild-dog-cat. It shows all weighted combinations, with corners representing pure classes and the interior points showing multi-class blends.

\section{Ablation Study}
\label{GAF:ablation_study}
The ablation study compares the full GAF objective, \(\mathcal{L}_{\text{GAF}}= \mathcal{L}_{\text{pair}}+\lambda_{\text{res}}\,\mathcal{L}_{\text{res}}+ \lambda_{\text{swap}}\,\mathcal{L}_{\text{swap}}\), against a pair-only variant trained with \(\mathcal{L}_{\text{GAF}}= \mathcal{L}_{\text{pair}}\) alone (i.e, removing residual and swap losses). 

\subsection{Effect of Swap and Residual Loss}
We report Pixel MSE, LPIPS(Alex)~\citep{zhang2018perceptual}, LPIPS(VGG), Latent MSE; lower is better for all metrics. The MSE captures the reconstruction error between the initial sample image and final sample image under a given random cyclic transport, while LPIPS captures perceptual similarity. Latent space metrics measure transport consistency before VAE decoding.
\begin{figure}[!htbp]
    \centering
    \includegraphics[width=0.7\linewidth]{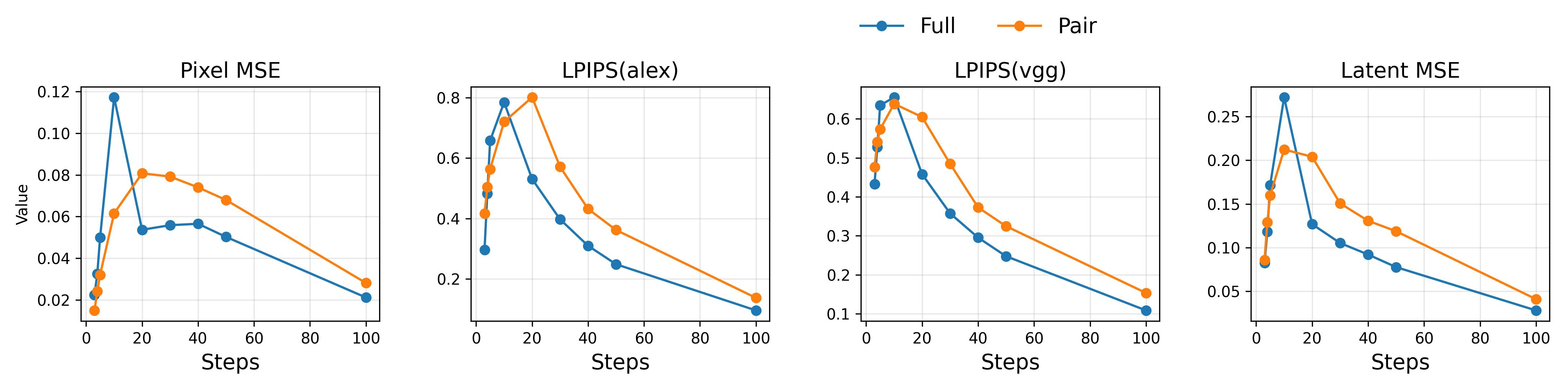}
    \caption{Effect of swap and residual losses under cyclic transport (\(z_y\rightarrow K_0\xrightarrow{J}K_1\xrightarrow{J}K_2\xrightarrow{J}K_0\)).}
    \label{fig:swap_res_loss_ablation_progress}
\end{figure}

As shown in Figure~\ref{fig:swap_res_loss_ablation_progress}, across the cyclic transport, the full objective yields consistently lower error rates and more stable multi-class transport at fewer forward steps, compared to only the pair loss. The error values of the full and pair objectives values converge as \(N\) approaches 100. In conclusion, the \(\mathcal{L}_{\text{res}}\) and \(\mathcal{L}_{\text{swap}}\) are not required for endpoint learning; instead, they primarily improve stability, efficiency, and sampling speed. 

\subsection{Image-to-Image Translation}
\label{img_to_img_translation}
\begin{figure}[!htbp]
    \centering
    \includegraphics[width=0.75\linewidth]{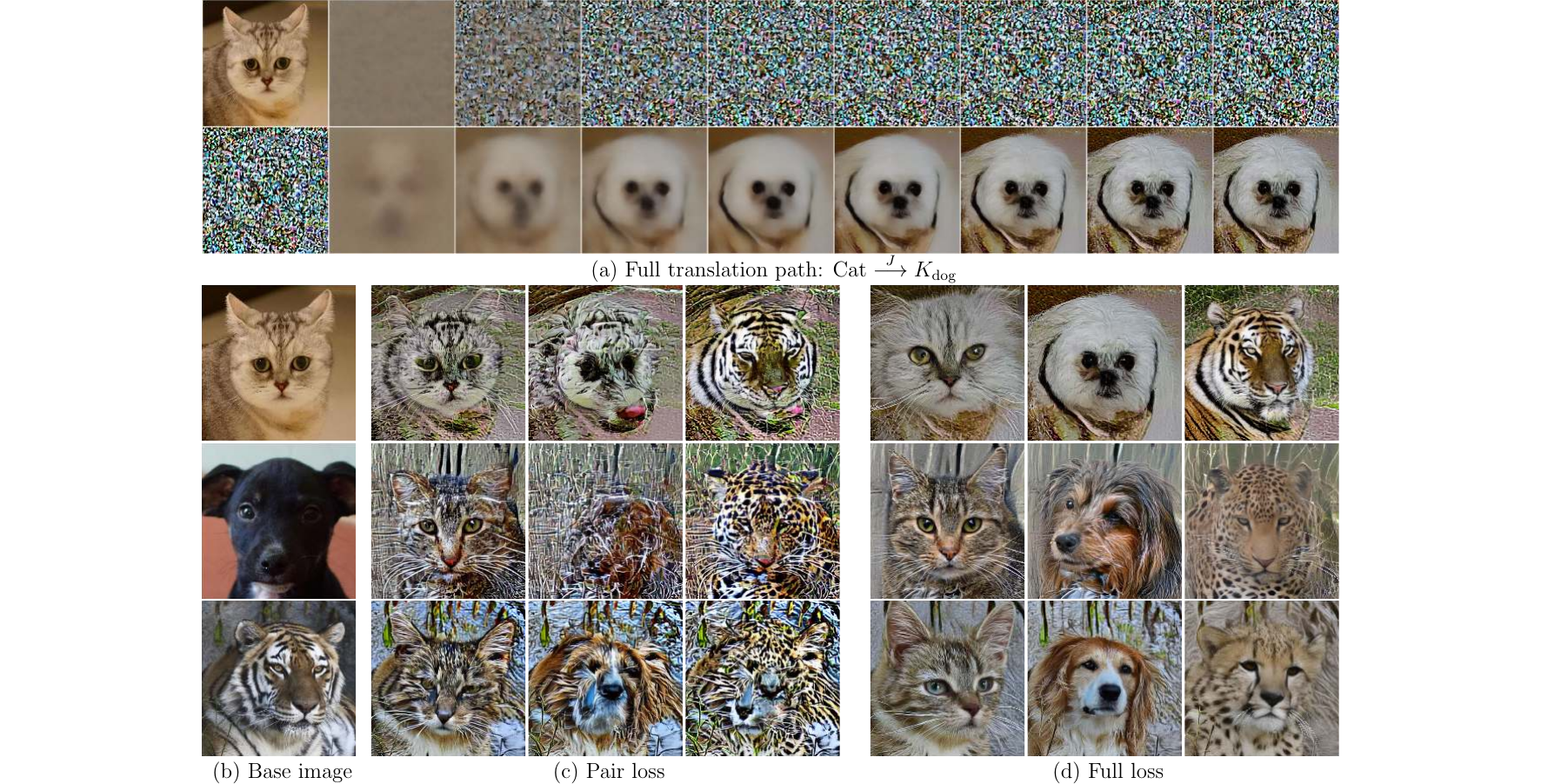}
    \caption{Image-to-Image translation (8-step IER). (a) Visualization of the full cross-modality translation path (\(K_{\text{cat}}\xrightarrow{J}K_{\text{dog}}\)). (b) Base images from the held-out AFHQ dataset. (c) Translations generated by GAF trained only with pair loss (\(\mathcal{L}_{\text{pair}}\)), which exhibit significant drift and local artifacts. (d) Translations utilizing the full loss objective, demonstrating that the residual and swap losses are crucial for stabilizing the latent space and preserving semantic structure.}
    \label{fig:zero_shot_translation}
\end{figure}

Figure~\ref{fig:zero_shot_translation}a shows an 8-step IER image-to-image translation using cross-modality transport between the cat and dog classes, as described in Section~\ref{GAF:cross_modality_transport}.  The full translation path shown in Figure~\ref{fig:zero_shot_translation}a is \(K_\text{cat}\rightarrow{J}\rightarrow{K_\text{dog}}\): we first invert a cat image back to noise and then generate an image conditioned on the dog class. The same translation procedure is applied for the other classes. The base images in Figure~\ref{fig:zero_shot_translation}b are from the AFHQ held-out dataset. The images generated with GAF trained using the \(\mathcal{L}_{\text{pair}}\)-only objective (Figure~\ref{fig:zero_shot_translation}c) show more drift and local artifacts and inconsistent multi-class generation, while the full loss objective (Figure~\ref{fig:zero_shot_translation}d) preserves identity and semantic structure more reliably under chained transport. This indicates that the residual and the swap losses are important for stabilization of the latent space.

\subsection{K-only Endpoint Ablation}
We additionally train a GAF variant containing only the \(K\) endpoint, without CFG. The \(K\)-only GAF achieves an FID of 8.08 when trained from a pretrained DiT model, which is comparable to the full \textit{retrunked} GAF, with an FID of \(7.51\). 

\section{Comparison with Rectified Flow}
\label{GAF:rectified_flow_comparison}
GAF learns endpoint operators \(J\) and \(K\) and samples via IER, while Rectified Flow~\citep{liu2022flow} learns a velocity field \(v_{\theta}(\mathbf{x_t},t)\) and requires ODE integration. GAF has an equivalent velocity field \(v=K-J\). Under ODE sampling, both models exhibit comparable compositional capacity. However, GAF can use both IER and ODE sampling, even within a single generation. 
\begin{figure}[!htbp]
  \centering
  \includegraphics[width=0.85\linewidth]{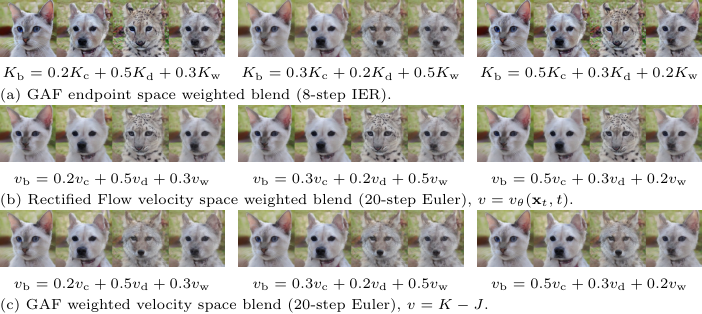}
  \caption{GAF comparison with Rectified Flow under identical compositions. The first three images in the grid are the base images, while the fourth is the weighted blend.}
  \label{fig:rect_comparison}
\end{figure}

Figures \ref{fig:rect_comparison}a, \ref{fig:rect_comparison}b, and \ref{fig:rect_comparison}c illustrate endpoint composition with IER and its qualitative comparison to two ODE-based samplers, respectively. IER operates directly on the predicted coordinates without solving an ODE, which is a capability Rectified Flow's velocity-only formalism cannot support. Because GAF exposes its two endpoints, \(J\) and \(K\), it also enables Transport Algebra operations directly in endpoint space and native bidirectional sampling, using \(K\) for forward generation and \(J\) for reverse inversion. Neither capability is present in a velocity-only formalism. For example, with GAF, one can compose endpoints as \(0.5K_{\text{dog}}+0.3K_{\text{cat}}+0.2K_{\text{wild}}\) to generate a sample with mixed dog, cat, and wild features. Rectified Flow approximates this by constructing a composite velocity field but must route through ODE integration. In contrast, GAF supports endpoint composition via IER, velocity composition via ODEs, and mid-trajectory switching between the two. 

\section{Limitations}
\label{sec:limitations}
While the computational overhead of the linear predictors is negligible, they add additional memory cost  for multi-class training compared to models such as Rectified Flow. For example, 1000-class ImageNet requires 1000 linear predictors. The swap loss requires an additional forward pass at the swapped input, doubling the computational cost required to train the model. Although the IER sampler converges in 5-8 steps, faster than Euler, it does not match the visual quality (measured in FID) of models designed for one-step generation (e.g, MeanFlow\citep{geng2026mean}). Closing this quality gap while retaining IER's fast convergence remains an important direction for future work.

\section{Broader Impacts}
\label{sec:broader_impacts}
Our work on GAF shares similar impact associated with current generative models, such as diffusion and flow-based models. GAF can be used for positive and negative purposes, however, within the scope of this work we do not foresee misuse risks beyond current generative image models.


\end{document}